\documentclass[sigconf, authorversion, nonacm]{acmart}

\usepackage{booktabs} %

\usepackage[ruled]{algorithm2e} %
\usepackage{float}

\SetAlFnt{\small}
\SetAlCapFnt{\small}
\SetAlCapNameFnt{\small}
\SetAlCapHSkip{0pt}

\acmJournal{TOG}

\usepackage{amsmath}
\usepackage{comment}
\usepackage{multirow,bigdelim}
\usepackage{lipsum}
\usepackage{array}
\usepackage{wrapfig}

\usepackage{adjustbox}
\usepackage[percent]{overpic}
\usepackage{makecell}
\usepackage[normalem]{ulem}
\usepackage{blindtext}
\usepackage{xcolor}
\usepackage{soul}
\usepackage{cleveref}
\usepackage{amsfonts}
\usepackage{subcaption}

\makeatletter
\newcommand{\settitle}{\@maketitle}
\makeatother

\newcolumntype{C}[1]{>{\centering\let\newline\\\arraybackslash\hspace{0pt}}m{#1}}

\newif\ifdraft
\drafttrue

\ifdraft
\definecolor{darkpink}{rgb}{0.561, 0.282, 0.427}
\definecolor{darkturquoise}{rgb}{0., 0.81, 0.822}

\newcommand{\dcc}[1]{{\color{red}[\textbf{DC:} #1]}}
\newcommand{\rgc}[1]{{\color{purple}[\textbf{RG:} #1]}}
\newcommand{\opc}[1]{{\color{blue}[\textbf{OP:} #1]}}

\newcommand{\abc}[1]{{\color{green}[\textbf{AB:} #1]}}

\newcommand{\drop}[1]{}

\else
\newcommand{\dcc}[1]{}
\newcommand{\rgc}[1]{}
\newcommand{\opc}[1]{}
\newcommand{\gcc}[1]{}
\newcommand{\hmc}[1]{}
\newcommand{\abc}[1]{}

\fi

\def\naive{na\"{\i}ve\xspace}

\newcommand{\wq}{${W}_q$\xspace}
\newcommand{\wk}{${W}_k$\xspace}
\newcommand{\wv}{${W}_v$\xspace}

\makeatletter
\DeclareRobustCommand\onedot{\futurelet\@let@token\@onedot}
\def\@onedot{\ifx\@let@token.\else.\null\fi\xspace}

\def\eg{\emph{e.g}\onedot}

\def\ie{\emph{i.e}\onedot}

\def\etal{\emph{et al}\onedot}

\def\pholdercolor{{\color{blue}$S_*$}}

\makeatother

\usepackage[bottom]{footmisc}
\usepackage{arydshln}

\makeatletter
\def\blfootnote{\xdef\@thefnmark{}\@footnotetext}
\makeatother

\begin{document}
\title{Encoder-based Domain Tuning for Fast Personalization of Text-to-Image Models}

\author{Rinon Gal}
\affiliation{%
 \institution{Tel Aviv University, NVIDIA}
 \city{Tel Aviv}
 \country{Israel}}
 \authornote{work was done during an internship at NVIDIA}

\author{Moab Arar}
\affiliation{%
 \institution{Tel Aviv University}
 \city{Tel Aviv}
 \country{Israel}}
 
\author{Yuval Atzmon}
\affiliation{%
 \institution{NVIDIA}
 \city{Tel Aviv}
 \country{Israel}}
 
\author{Amit H. Bermano}
\affiliation{%
 \institution{Tel Aviv University}
 \city{Tel Aviv}
 \country{Israel}}

\author{Gal Chechik}
\affiliation{%
 \institution{NVIDIA}
 \city{Tel Aviv}
 \country{Israel}}
 
\author{Daniel Cohen-Or}
\affiliation{%
 \institution{Tel Aviv University}
 \city{Tel Aviv}
 \country{Israel}}

\begin{abstract}

Text-to-image personalization aims to teach a pre-trained diffusion model to reason about novel, user provided concepts, embedding them into new scenes guided by natural language prompts. However, current personalization approaches struggle with lengthy training times, high storage requirements or loss of identity. To overcome these limitations, we propose an encoder-based \textit{domain-tuning} approach. Our key insight is that by \textit{underfitting} on a large set of concepts from a given domain, we can improve generalization and create a model that is more amenable to quickly adding novel concepts from the same domain. Specifically, we employ two components: First, an encoder that takes as an input a single image of a target concept from a given domain, \eg a specific face, and learns to map it into a word-embedding representing the concept. Second, a set of regularized weight-offsets for the text-to-image model that learn how to effectively ingest additional concepts. Together, these components are used to guide the learning of unseen concepts, allowing us to personalize a model using only a single image and as few as $5$ training steps --- accelerating personalization from dozens of minutes to \textit{seconds}, while preserving quality.

Code and trained encoders will be available at {\color{blue}{\href{https://tuning-encoder.github.io/}{our project page.}}}

\end{abstract}
\begin{teaserfigure}
    \setlength{\abovecaptionskip}{5.5pt}
    \setlength{\belowcaptionskip}{1.5pt}
    \setlength{\tabcolsep}{0.55pt}
    \centering
     \includegraphics[height=6cm]{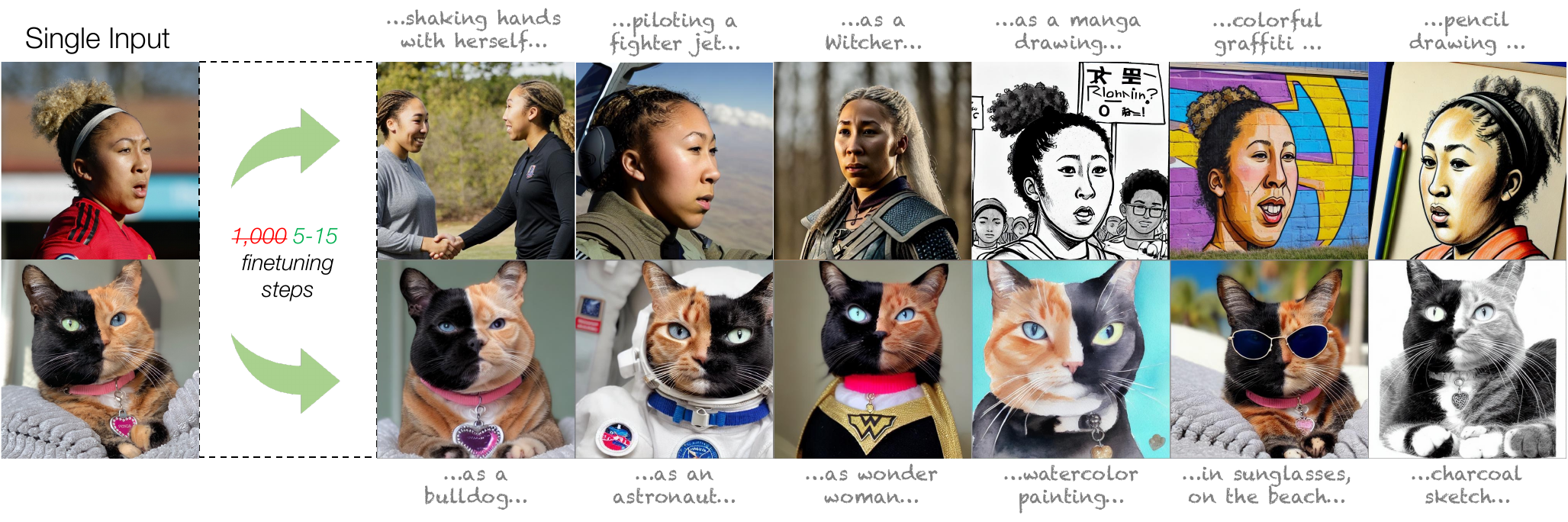}
    \caption{Our encoder-based method enables quick personalization of text-to-image models, teaching them novel concepts in seconds rather than dozens of minutes. The personalized model can then be used to create new images of the concept using natural language prompts.}
    \label{fig:teaser}
\end{teaserfigure}
\maketitle

\section{Introduction}

The ability to personalize large-scale text-to-image models~\citep{ramesh2022hierarchical,rombach2021highresolutionLDM,saharia2022photorealistic} has revolutionized content creation. By injecting an understanding of a new concept into pre-trained models, a user can leverage it in new prompts, thereby inserting a subject into new scenes, or invoking a unique artistic style with a single word. 

However, current personalization methods~\citep{gal2022image,ruiz2022dreambooth} are difficult to scale, with each new concept requiring fine-tuning sessions lasting dozens of minutes or even hours on a top-end GPU. Moreover, when tuning the entire model, the resulting checkpoints are typically several GB in size, incurring non-negligible storage and serving costs. Finally, to prevent the model from overfitting spurious image details such as the background, these models typically require painstaking collection of multiple images with varied backgrounds and poses.

We propose to tackle these challenges through a novel \textit{domain-tuning} approach, where the text-to-image model is taught how to personalize well to new concepts from a given domain. For example, we would like to tune the model on a dataset of cats (the domain), such that it can more easily be personalized to individual, unseen cats (the concepts). Our approach is based on the premise that highly regularized models are well suited to learning some averaged behaviour. We thus propose to concurrently personalize a network on a large collection of concepts from a single domain, while limiting its degrees of freedom. Instead of learning the details of individual concepts, this network then learns a more generalized set of weights that lies close to each of them individually. Our approach is thus similar in spirit to Meta Learning~\citep{finn2017model}, and specifically the joint-training scenario outlined by \citet{nichol2018reptile}.

To implement our approach, we design two components: A restricted set of weights to tune, and an efficient way to invert a large number of concepts. For the weights, we investigate models tuned with prior approaches and propose to modulate the projection matrices in the denoising network's attention mechanism. Rather than tuning the weights directly, we employ a learned constant followed by a set of fully connected layers to transform it into an offset for the model's weights. Learning these changes through a network serves to restrict both the rank of the learned offsets and to provide a smoothness prior~\citep{rahaman2019spectral}.

For efficient inversion, we employ an encoder -- a neural network tasked with quickly mapping a given concept image into a word embedding that approximately represents it. We draw inspiration from the literature on GAN inversion~\citep{xia2021gan,bermano2022state} and propose an iterative-refinement~\citep{alaluf2021restyle} scheme. Here, instead of predicting an embedding representation for the concept in a single forward-pass, we couple the encoder to the denoising process and predict a novel embedding for each time step. Moreover, through this iterative approach, the encoder can observe both the target image and the current noisy sample at each step. This allows it to correct for mistakes during the synthesis process.

The encoder and weight offsets are jointly pretrained on a large dataset from a single given domain, such as FFHQ~\citep{karras2019style}, LSUN Cat~\citep{yu2015lsun}, or WikiArt~\citep{saleh2015large}. To learn a new, specific concept at inference time, we fine-tune both components and the diffusion model on a single image portraying the personal concept. Our approach thus serves three goals: (1) a strong initialization to the tuning process for both the model and the word embedding, (2) a means of allowing the network to correct for mistakes during the iterative denoising process, and (3) a domain prior that helps identify the target concept even from a single image. Together, these allow us to tune a model for a specific concept with \textit{a single image} and as few as $5$ training iterations --- roughly 11 seconds of training on a single NVIDIA A100 GPU, or $\times 60-140 $ faster than previous personalization approaches. Importantly, as tuning now takes a number of steps comparable to the synthesis process, it can be used at inference time to enable one-shot personalization, without requiring a new model for every new identity.

We compare our approach with prior personalization baselines and demonstrate that our method can synthesize appealing results, with fewer images, and with a fraction of the tuning time.

\section{Related work}

\paragraph{\textbf{Text-guided synthesis.}} Early text-to-image models employed a Generative Adversarial Network (GAN)-based architecture~\citep{goodfellow2014generative} trained on large collections of paired image-caption data~\citep{zhu2019dm,tao2020df,xu2018attngan,zhang2021cross,ye2021improving}. However, GANs are prone to mode collapse and are difficult to train at scale~\citep{BrockDS19,HeuselRUNH17}. Motivated by the scaling success of language models, auto-regressive models~\citep{ramesh2021zero,yu2022scaling, MakeASceneGafni} treated images as word sequences in a discrete latent space~\citep{VQVAEOord, VQGANEsser}. There, text guidance could be used by conditioning the generation on text-prefix, or through test-time optimization using text-to-image similarity models~\cite{radford2021learning,katherine2021vqganclip}. Recently, diffusion models~\cite{ho2020denoisingDDPM, nichol2021improvedDDPM, dhariwal2021diffusionBeatsGAN} have taken the front in image generation. They led to a remarkable headway in text-to-image synthesis, achieving unprecedented diversity and fidelity~\citep{ramesh2022hierarchical,saharia2022photorealistic,nichol2021glide,rombach2021highresolutionLDM,balaji2022ediffi}. Our approach aims to leverage such pre-trained text-to-image diffusion models, and teach them to reason about personalized concepts.

\paragraph{\textbf{Inversion.}} Image inversion refers to the task of finding a latent code that can be fed into a generator to reconstruct a given target image~\citep{zhu2016generative,xia2021gan}. This process typically involves direct optimization of the latent on a single image~\citep{abdal2019image2stylegan,abdal2020image2stylegan++,zhu2020improved,gu2020image,parmar2022spatially} or training a neural network to predict such a code directly. Such a network is typically referred to as an encoder~\citep{richardson2020encoding,zhu2020domain,pidhorskyi2020adversarial,tov2021designing}, and is trained on large datasets, allowing it to generalize to new targets. When the generator's latent spaces exhibit strong semantics, these codes can be manipulated in order to edit the target image~\citep{shen2020interpreting, patashnik2021styleclip,gal2021stylegan} or used for regression tasks~\citep{xu2021generative, nitzan2021large}. 

With diffusion models, inversion often refers to the task of finding an initial noise sample that can be denoised into a given target~\citep{dhariwal2021diffusionBeatsGAN,ramesh2022hierarchical}. Unfortunately, such methods do not lend themselves well to downstream editing, leading to a loss of identity when the conditioning code is modified~\cite{gal2022image,mokady2022null}. More recently, inversion has been used in the context of text-to-image synthesis to describe the task of finding a latent code that can be used to synthesize novel images of a given concept~\citep{gal2022image}, a process also referred to as \textit{personalization}. Our method falls into the latter category.

\paragraph{\textbf{Personalization.}} 
Personalization methods adapt a given model to a unique individual or group by leveraging data specific to the target user. These methods have been used in various applications, such as recommendation systems~\citep{benhamdi2017personalized,fernando2018artwork,martinez2009s,cho2002personalized}, federated learning~\citep{mansour2020three,jiang2019improving,fallah2020personalized,shamsian2021personalized}, and, more recently, in computer vision and graphics~\citep{semantic2019bau,roich2021pivotal,alaluf2021hyperstyle,dinh2022hyperinverter,cao2022authentic,nitzan2022mystyle,cohen2022my,Agrawal2021KnownUL}. In the context of text-to-image diffusion models, prior work aims to teach a pre-trained model to synthesize novel images of a specific target concept, guided by natural language prompts. Current personalization methods either optimize a set of text embeddings to describe the concept~\citep{gal2022image} or tune the denoising network to tie a rarely-used word-embedding to the new concept~\citep{ruiz2022dreambooth}. However, both approaches require multiple images of the target concept, and employ lengthy training sessions, lasting dozens of minutes on high-end GPUs. Moreover, the model-tuning approach requires storing several gigabytes of data for each new concept - making large-scale personalization a costly endeavour. Our work brings the encoder-based approach to the realm of personalization, reducing training times by orders of magnitude, eliminating the need to store models, and allowing for model personalization using only a single image.

\section{Method}
\begin{figure*}
    \centering
    \setlength{\abovecaptionskip}{8.5pt}
    \setlength{\belowcaptionskip}{-6.5pt}
    \setlength{\tabcolsep}{0.55pt}

    \includegraphics[width=0.83\textwidth]{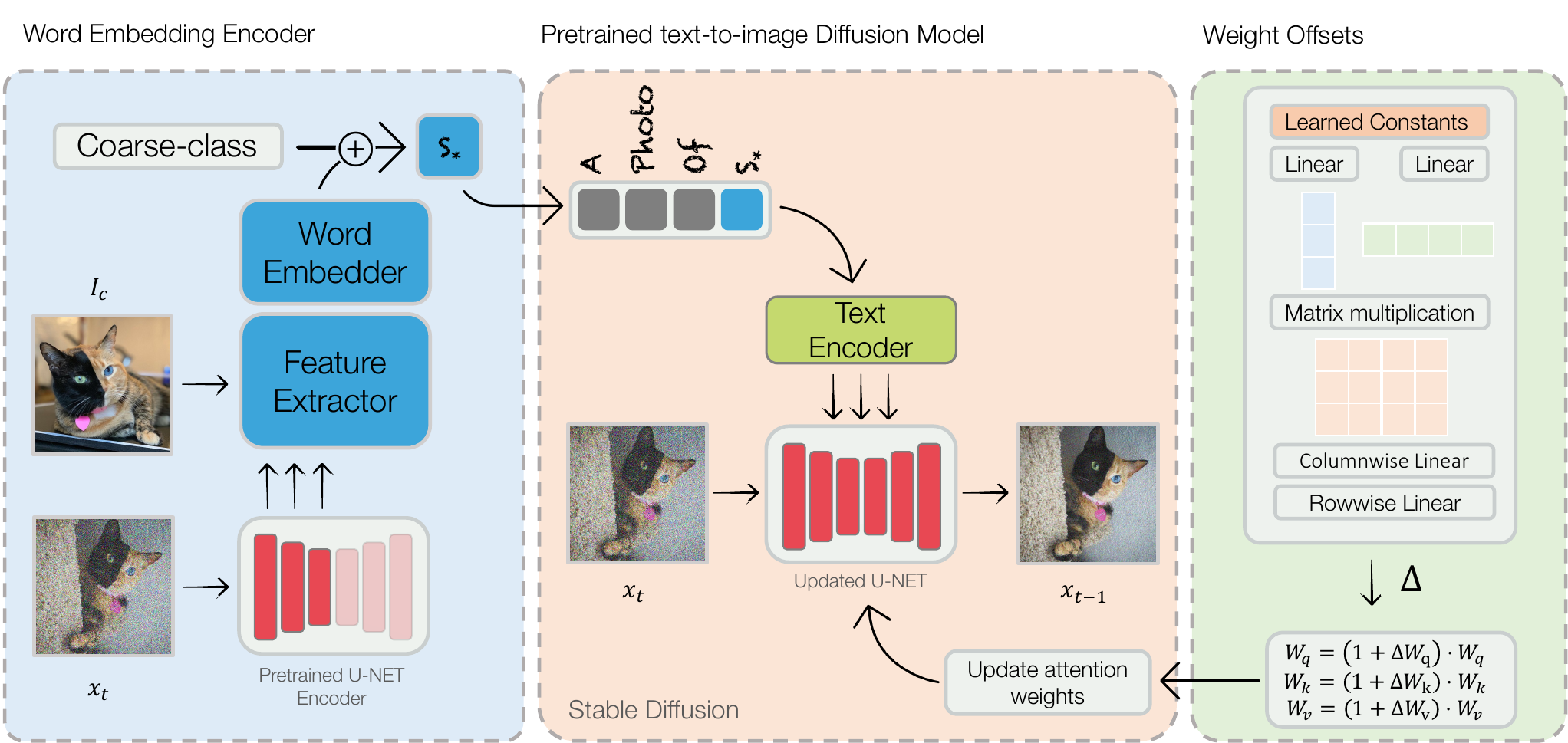}
    \caption{\textbf{Method overview.} Our method efficiently adapts pre-trained diffusion models (middle) to handle novel concepts. We train two components, a word-embedding encoder (left) and weight offsets (right). The encoder predicts a new code in the diffusion model's embedding space which best describes the input concept. In addition, we optimize learned weight offsets to specialize the text-to-image model to the target domain (\eg cats). Our offsets are learned constants, regularized through a neural network. During training, we use $x_0 = I_c$. Here, we show inference, where $x_T$ is sampled from the Gaussian prior as usual and hence $x_t$ is no longer a noisy version of $I_c$.}
    \label{fig:arch-fig}
\end{figure*}
Our goal is to design a method for efficient injection of new concepts into a pre-trained text-to-image diffusion model. To do so, we propose a \textit{domain-tuning} approach where the model is taught how to personalize well to new concepts from a given domain. To achieve this goal, we train an encoder for efficient inversion of concepts into the diffusion model, and a set of weight-offsets that modify the model so that it can be quickly tuned for novel concepts.

In the following section, we outline the task of encoder inversion, our design choices when creating such an encoder, and the motivations behind them. Then, we discuss our approach for selecting 
a subset of weights that is expressive enough to enable personalization, yet restrictive enough to prevent the model from overfitting to the training data. Finally, we discuss our tuning approach and additional tools that can improve the results.

\vspace{-3pt}
\subsection{Inversion and encoder design}
\textit{Inversion encoders} are neural networks that take an input image $I_c$ representing a specific concept and predict some latent code $z_c$ such that feeding the code back into the generator $G$ will result in a new image of the concept, \ie, $I' = G\left(z_c\right) = G\left(E\left(I_c\right)\right) \sim I_c$.

To train such an inversion encoder, we must first choose a suitable latent space in which concepts will be represented. In the case of text-to-image diffusion models, a possible candidate is the word embedding space used in Textual Inversion~\citep{gal2022image}. However, as Gal~\etal demonstrate, this space exhibits a trade-off between reconstruction and editability. This is because more accurate concept representations typically reside far from the real word embeddings, leading to poorer performance when using them in novel prompts. \citet{roich2021pivotal} tackled a similar hurdle for StyleGAN inversion. They proposed a two-step solution which consists of approximate-inversion followed by model tuning. The intuition here is that the initial inversion can be constrained to an editable region of the latent space, at the cost of providing only an approximate match for the concept. The generator can then be briefly tuned to shift the content in this region of the latent space, so that the approximate reconstruction becomes more accurate. Since this change is more localized, it minimizes the loss of the network's prior knowledge.

We aim to employ a similar tuning approach here and, therefore, wish the encoder to invert new concepts into an editable region of the word-embedding space, even at the cost of accuracy. Since we train a domain-specific encoder, we elect to maintain editability by constraining our predicted embeddings to reside near the word-embedding of the domain's coarse 
descriptor (\eg ``face", ``cat" or ``art"). Our concept-specific embedding $e_c$ is thus given by:
\begin{equation}
    \epsilon_c = \epsilon_{domain} + s \cdot E \left(I_c\right) ,
\end{equation}
where $E$ is our encoder, $\epsilon_{domain}$ is the pre-trained model's embedding for the domain's coarse descriptor, and $s$ is a scaling factor which we empirically set to $0.1$

We additionally constrain the encoder's prediction through a regularization penalty term: 
\begin{equation}
    \mathcal{L}_{reg} = ||E\left(I_c\right)||_2^2.
\end{equation}\label{eq:emb_regularization}

\vspace{-13pt}
\paragraph{\textbf{Encoder architecture}} 
We design our encoder as a set of feature-refinement blocks built on top of a pre-trained OpenCLIP~\citep{ilharco_gabriel_2021_5143773} ViT-H/14~\citep{dosovitskiy2020vit} feature-extraction backbone. Specifically, we extract the features of the [CLS] token of each 2nd CLIP layer as an hierarchical feature representation~\citep{tumanyan2022splicing,vinker2022clipascene}. Each such feature vector is fed through a linear layer, followed by average-pooling over the hierarchy and LeakyReLU activation. These features are then fed into a final linear layer which predicts the embedding offset, $E \left(I_c\right)$. 

\paragraph{\textbf{Iterative refinement}}

In the GAN literature, \citet{alaluf2021restyle} proposed an iterative inversion scheme. There, an encoder first predicts an initial latent $z_0$ for a given target image $I_t$. This code is fed into a GAN $G$ to generate an initial reconstruction $I'_0 = G\left(z_0\right)$. The encoder then receives the pair $\{I_t, I'_0\}$ and is tasked with reasoning over any discrepancies and predicting a refined latent $z_1$ that produces a better reconstruction, $I'_1 = G\left(z_1\right)$. This process continues in an iterative fashion, leading to increased reconstruction quality. 

Diffusion models already employ an iterative denoising process, and therefore lend themselves well to such an
approach. However, in contrast to GAN inversion, the denoising process does not provide us with clean reconstructions to use as an input to the encoder for the next iteration. Moreover, popular diffusion models operate in a latent domain~\citep{rombach2021highresolutionLDM}, and the codes they produce cannot be \naive{ly} fed into our feature extraction backbone. Using such a model's decoder to map the latents back to the image domain incurs a significant cost in both memory and time. 
Instead, we propose that the pre-trained diffusion model \textit{already contains} a feature extraction network that can reason over noisy latents, the denoiser's U-net itself.
Therefore, we feed the noisy image into the U-net encoder and extract the pooled features from each of its blocks. These features are then concatenated with the hierarchical features extracted from the concept image through the CLIP backbone, before being passed to the rest of the encoder. See \cref{fig:arch-fig} for an illustration of this process.

\subsection{Weight offsets}

Our approach requires us to determine a subset of model parameters that is expressive enough to allow for downstream personalization, yet restrictive enough that tuning it alone will not shift the entire generator's domain. To do so, we first examine a set of 50 fully tuned networks taken from the HuggingFace concept library~\citep{huggingface2022concept}. These range from specific objects to more abstract concepts such as artistic styles. To identify which layers underwent the most significant change during tuning, we begin by calculating the unsigned distance between each fine-tuned model's weights and those of the original. Each layer is assigned an importance score by taking the mean distance over its parameters and normalizing it by the mean value of parameters within the layer. Finally, we average each layer's score between all the tuned models and rank the layers according to their importance score. In \cref{tab:layer_importance} we list the importance score breakdowns according to different types of layers. 

We observe that the cross- and self-attention layers have higher scores, indicating that they play a crucial part in the tuning effort, and thus focus on modulating the weights of these layers. Specifically, we modify the three attention projection matrices - \wq, \wk and \wv.  We note that a similar study was conducted in the concurrent work of \citet{kumari2022multi}. Their conclusions are largely the same. However, they opted to focus their tuning on a smaller subset of layers.
\begin{table}[hbt]\setlength{\tabcolsep}{3pt}
\vspace{-3pt}
\setlength{\abovecaptionskip}{6.5pt}
\small
\caption{Layer importance scores. Attention layers see more drastic changes during the tuning process, and so we focus on them. }\label{tab:layer_importance}

\centering 
\begin{tabular}{lclc} 
    \toprule
    \multicolumn{4}{c}{By Layer Location} \\ 
    \midrule 
    Location & Score $\left(\uparrow\right)$ & Location & Score $\left(\uparrow\right)$ \\
    \midrule 
    Cross-attention & 59.20 & Upsampling Blocks & 24.11 \\
    Self-Attention & 47.35 & Downsampling Blocks & 15.57 \\
    Other & 10.50 & Middle Block & 17.64 \\
    \bottomrule
    \toprule
    \multicolumn{4}{c}{By Attention Type} \\
    \midrule
    Cross-Attention & Score $\left(\uparrow\right)$ & Self-Attention & Score $\left(\uparrow\right)$ 
    \\
    \midrule
    $\mathrm{W_Q}$ & 140.18 & $\mathrm{W_Q}$ & 47.35 \\
    $\mathrm{W_K}$ & 62.93 & $\mathrm{W_K}$ & 82.41 \\
    $\mathrm{W_V}$ & 34.62 & $\mathrm{W_V}$ & 74.48 \\
    $\mathrm{W_O}$ & 30.78 & $\mathrm{W_O}$ & 16.26 \\
    \bottomrule
\end{tabular}
\vspace{-8pt}
\end{table}

However, our experiments indicate that this set of weights is still too permissive, leading to poor results at the concept-specific tuning stage. We thus propose to regularize the weight predictions by using a deep neural network as a prior~\citep{ulyanov2018deep}.
Let $\mathrm{W}_{\{q,k,v\}}$ be an attention projection matrix of size $MxN$, rather than predicting it directly, we propose to learn an initial parameter vector $v_0$ and four linear layers. The first two are used to project the vector into two components: $v_y \in \mathbb{R}^{Mx1}$ and $v_x \in \mathbb{R}^{1xN}$. These are multiplied to create an $MxN$ matrix. We further refine the result by applying a linear projection over the matrix rows, followed by a linear projection over the columns. This approach restricts the rank of the learned weights, and promotes a smoother, low-frequency result, preventing the network from overfitting to the concepts.  

Finally, rather than learning a new set of weights directly, we use the offset formulation of~\cite{alaluf2021hyperstyle}, i.e.:
\begin{equation}
    {W_{\{q,k,v\}}^{i}} = {W}_{\{q,k,v\},0}^{i} \cdot \left(1 + \Delta W_{\{q,k,v\}}^{i}\right) ,
\end{equation}
where ${W}_{*,0}^{i}$ are the initial weights of attention matrix $*$ at layer $i$ and $\Delta W_{*,i}$ are the learned offsets for the same layer.

\subsection{Pre-training}

We pre-train both the inversion encoder and the weight-offsets over a large image collection portraying our target domain. For faces we use a mix of both FFHQ~\citep{karras2019style} and CelebA-HQ~\citep{karras2017progressive}. For cats we use LSUN-Cat~\citep{yu2015lsun} and for artistic styles we use WikiArt~\citep{saleh2015large}.

Our loss is a mixture of the regularization loss of \cref{eq:emb_regularization} and the simple diffusion denoising loss~\citep{ho2020denoisingDDPM}:
\begin{equation}
    L_{Diffusion} := \mathbb{E}_{z, y, \epsilon \sim \mathcal{N}(0, 1), t }\Big[ \Vert \epsilon - \epsilon_\theta(z_{t},t,y) \Vert_{2}^{2}\Big] \, ,
    \label{eq:LDM_loss}
\end{equation}

\begin{equation}\label{eq:loss}
L = L_{Diffusion} + \lambda_r L_{reg} ,
\end{equation}
where $t$ is the time step, $z_t$ is an image or latent noised to time $t$, $\epsilon$ is the unscaled noise sample, and $\epsilon_\theta$ is the denoising network. For pre-training, we empirically set $\lambda_r$ to $0.01$.

\subsection{Inference-time Personalization}

As a final stage in the personalization process, we tune both of our components as well as the pre-trained diffusion model using a single image of the target concept and the same loss of \cref{eq:loss}. Importantly, even though the model is tuned with only a single image, we find it crucial to use a large batch size of 16 or more images. This is because the diffusion training process samples a different level of noise for each element in the batch. Hence, a large batch ensures that our model observes the concept across multiple time scales and can better adapt the iterative-refinement approach.

Finally, we find that for the human face domain, it is helpful to use an off-the-shelf face segmentation network~\cite{deng2019arcface} to mask the diffusion loss at this stage.

\subsection{Implementation details}
For a base text-to-image model, we employ Stable Diffusion~\citep{rombach2021highresolutionLDM}, the current state-of-the-art publicly available model. We pre-train our encoders and weight offsets using a base learning rate of $1e-6$ and a batch size of 16 on a single A100 GPU. The Stable Diffusion codebase scales learning rates by the batch size and number of GPUs, giving us an effective learning rate of $1.6e-5$.

Our face model was trained for $30,000$ steps, the cat model for $60,000$ steps, and the art model for $100,000$ steps. 

When fine-tuning for a specific concept, we set $\lambda_r = 0.1$ for the face domain and train for $15$ iterations. For the cat and art domains we set $\lambda_r = 1e-4$, the base learning rate to $3e-6$, and tune the model for $5$ iterations.

\begin{figure*}

    \centering
    \setlength{\belowcaptionskip}{-6pt}
    \setlength{\tabcolsep}{1.5pt}
    {\small
    \begin{tabular}{c c c c c c c c c }

        \multirow{2}{*}{Input} & Stable & \multicolumn{2}{c}{Textual Inversion} & \multicolumn{2}{c}{Dreambooth} & Textual Inversion & E4T (Ours) & \\

        & Diffusion & 1-image & 5-images & 1-image & 5-images & + Dreambooth & 1-image & \\

        \includegraphics[width=0.117\textwidth,height=0.117\textwidth]{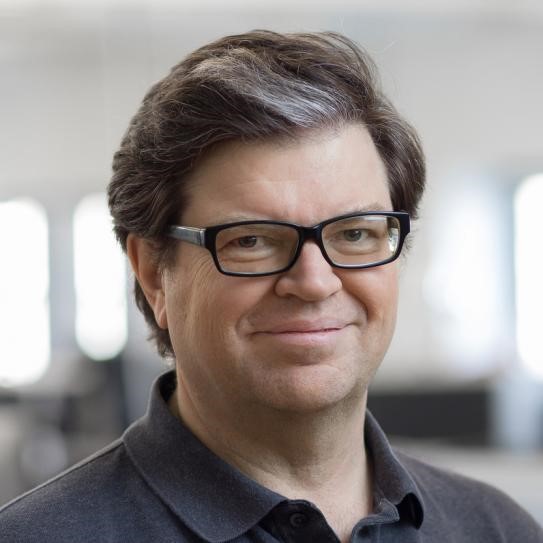} &
        \includegraphics[width=0.117\textwidth,height=0.117\textwidth]{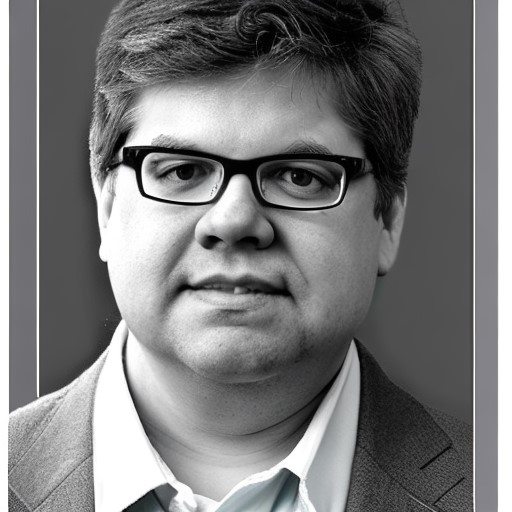} &
        \includegraphics[width=0.117\textwidth,height=0.117\textwidth]{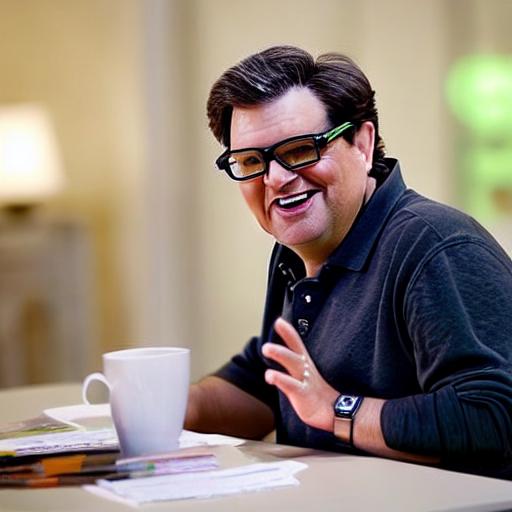} & %
        \includegraphics[width=0.117\textwidth,height=0.117\textwidth]{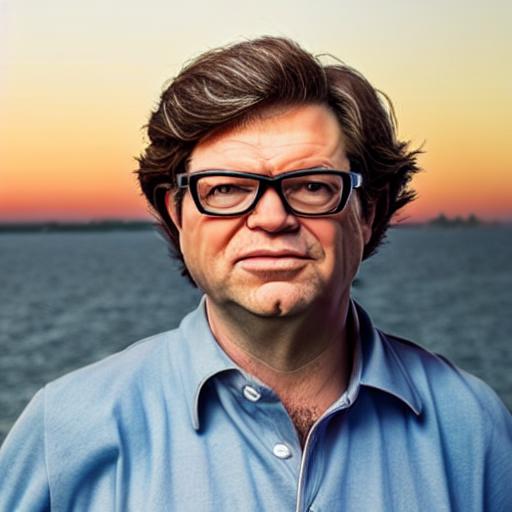} & %
        \includegraphics[width=0.117\textwidth,height=0.117\textwidth]{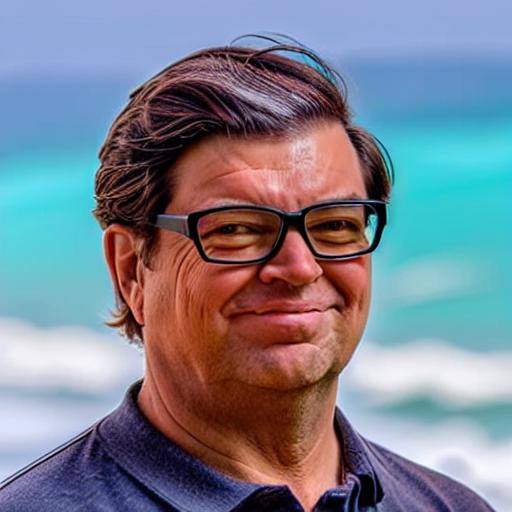} & %
        \includegraphics[width=0.117\textwidth,height=0.117\textwidth]{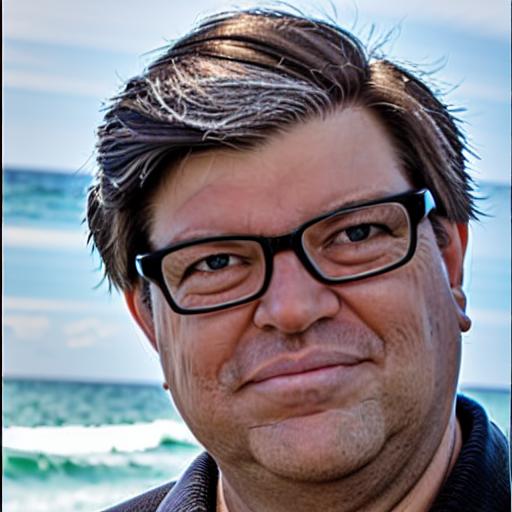}& %
        \includegraphics[width=0.117\textwidth,height=0.117\textwidth]{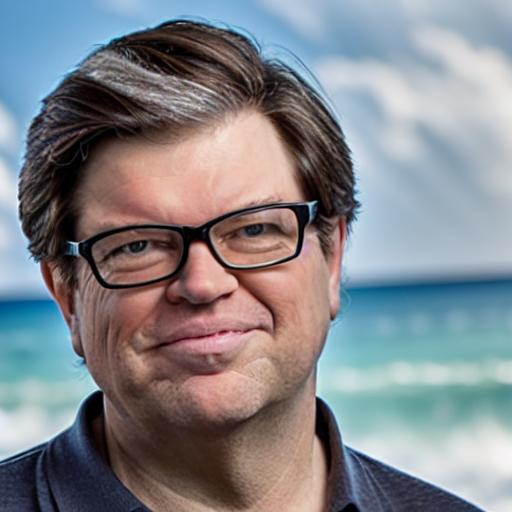} &
        \includegraphics[width=0.117\textwidth,height=0.117\textwidth]{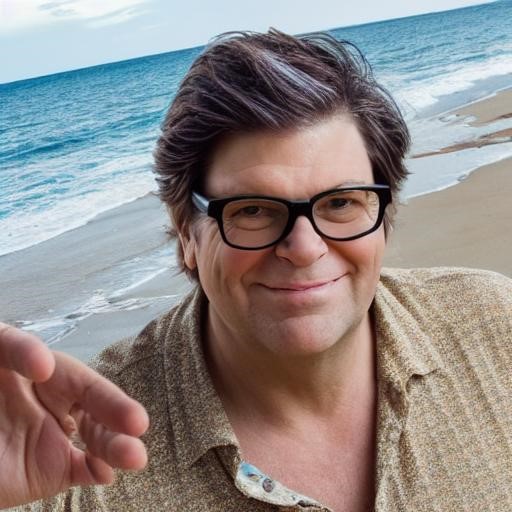} & \raisebox{0.048\textwidth}{\rotatebox[origin=t]{-90}{\scalebox{0.9}{\begin{tabular}{c@{}c@{}c@{}} \pholdercolor{} on the beach\end{tabular}}}}
        \\

        \includegraphics[width=0.117\textwidth,height=0.117\textwidth]{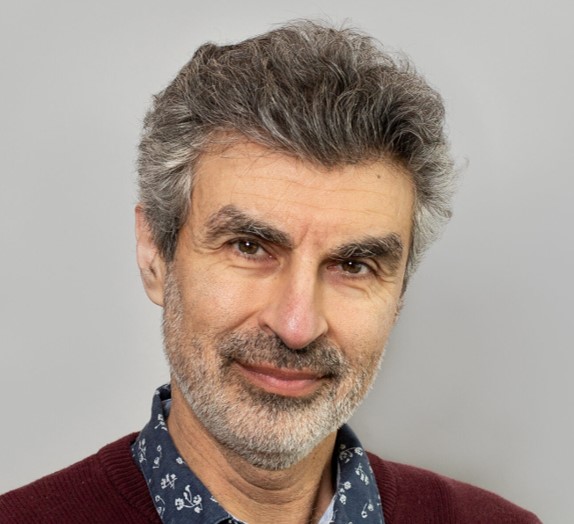} &
        \includegraphics[width=0.117\textwidth,height=0.117\textwidth]{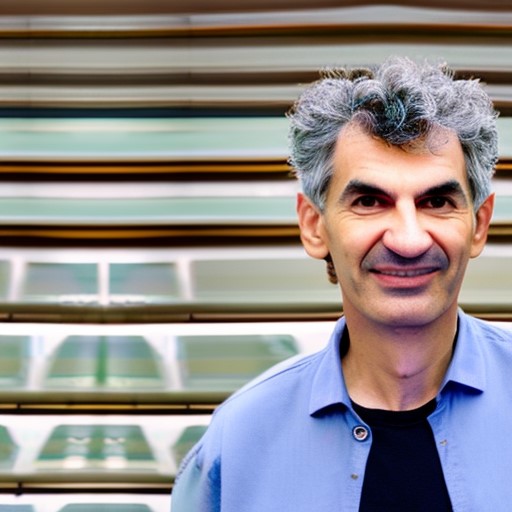} &
        \includegraphics[width=0.117\textwidth,height=0.117\textwidth]{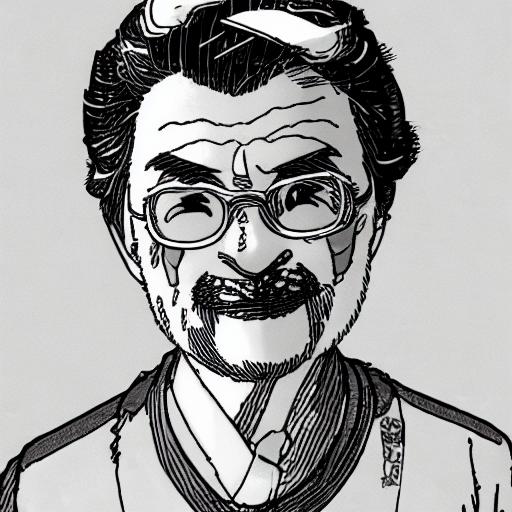} & %
        \includegraphics[width=0.117\textwidth,height=0.117\textwidth]{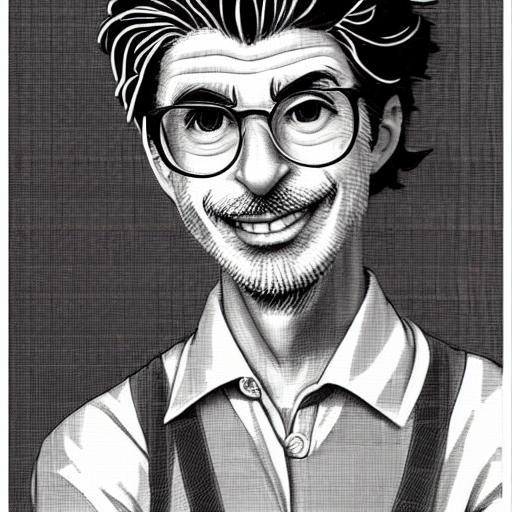} & %
        \includegraphics[width=0.117\textwidth,height=0.117\textwidth]{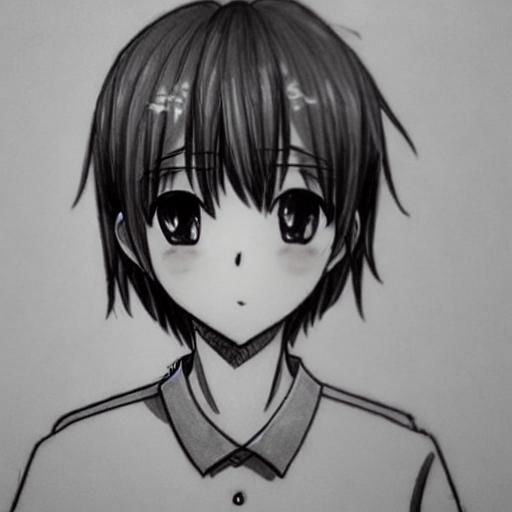} & %
        \includegraphics[width=0.117\textwidth,height=0.117\textwidth]{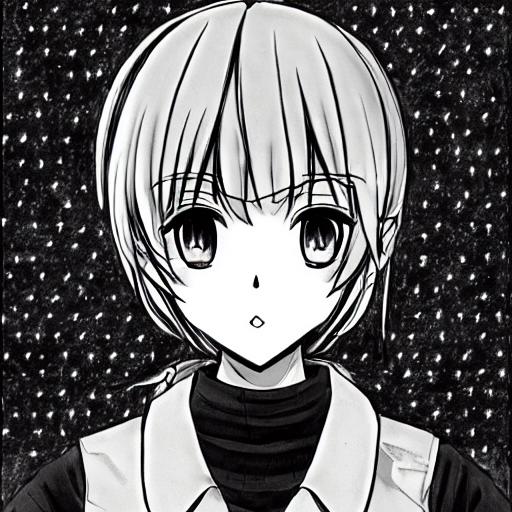}& %
        \includegraphics[width=0.117\textwidth,height=0.117\textwidth]{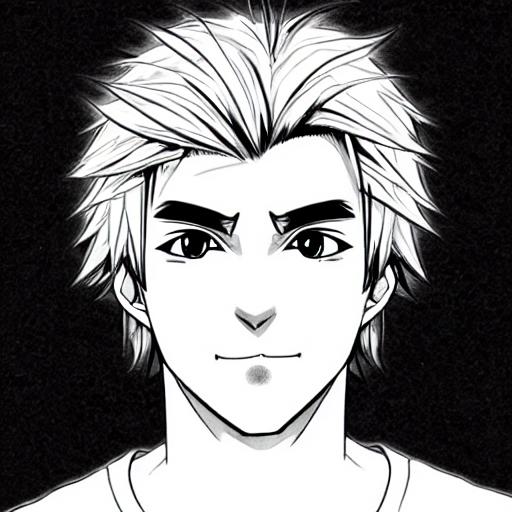} &
        \includegraphics[width=0.117\textwidth,height=0.117\textwidth]{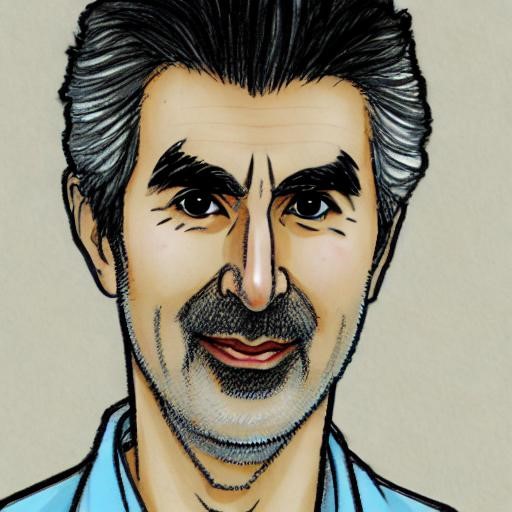} & \raisebox{0.035\textwidth}{\rotatebox[origin=t]{-90}{\scalebox{0.9}{\begin{tabular}{c@{}c@{}c@{}} Manga drawing \\ of \pholdercolor{}\end{tabular}}}} \\

        \includegraphics[width=0.117\textwidth,height=0.117\textwidth]{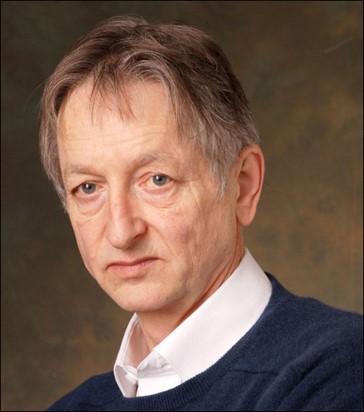} &
        \includegraphics[width=0.117\textwidth,height=0.117\textwidth]{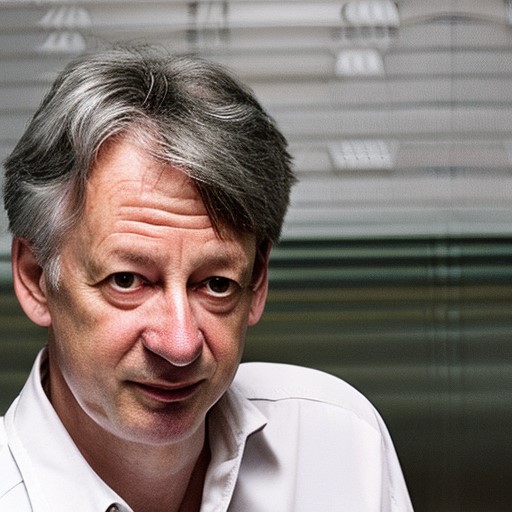} &
        \includegraphics[width=0.117\textwidth,height=0.117\textwidth]{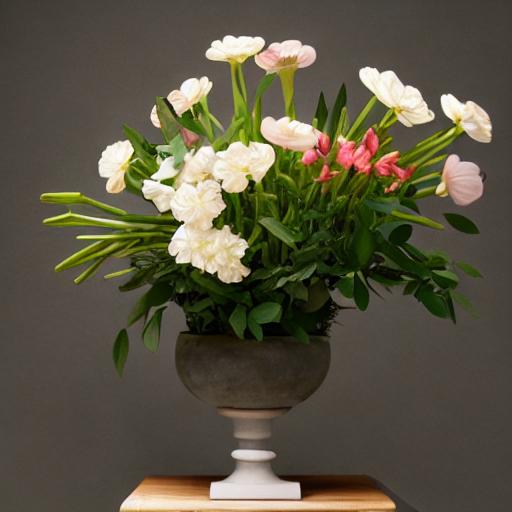} & %
        \includegraphics[width=0.117\textwidth,height=0.117\textwidth]{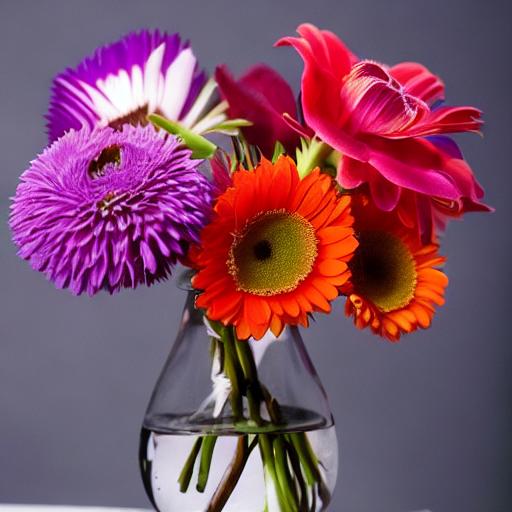} & %
        \includegraphics[width=0.117\textwidth,height=0.117\textwidth]{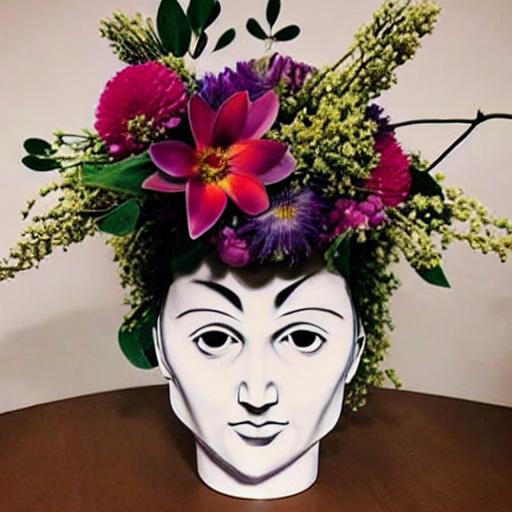} & %
        \includegraphics[width=0.117\textwidth,height=0.117\textwidth]{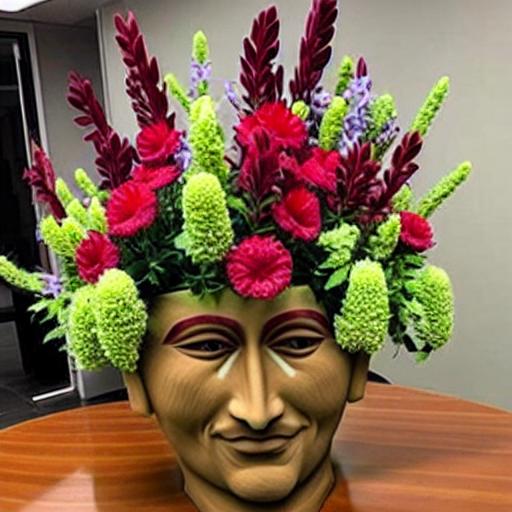}& %
        \includegraphics[width=0.117\textwidth,height=0.117\textwidth]{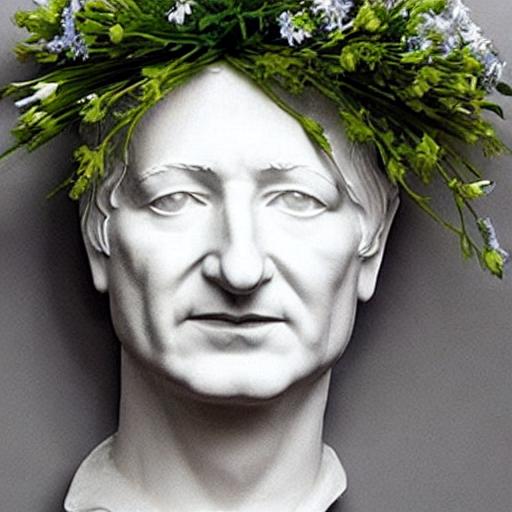} &
        \includegraphics[width=0.117\textwidth,height=0.117\textwidth]{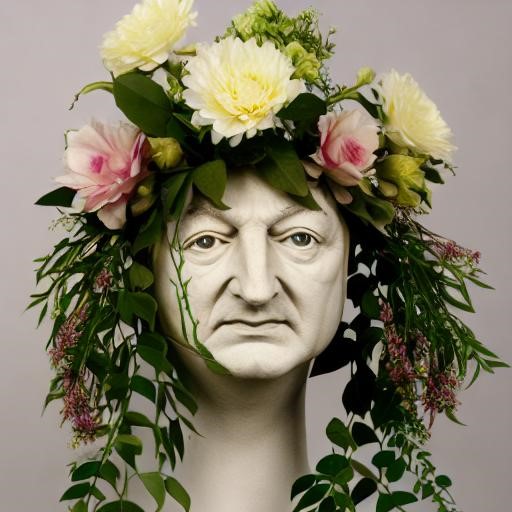} & \raisebox{0.035\textwidth}{\rotatebox[origin=t]{-90}{\scalebox{0.9}{\begin{tabular}{c@{}c@{}c@{}} \pholdercolor{} flower \\ arrangement \end{tabular}}}} \\

        \includegraphics[width=0.117\textwidth,height=0.117\textwidth]{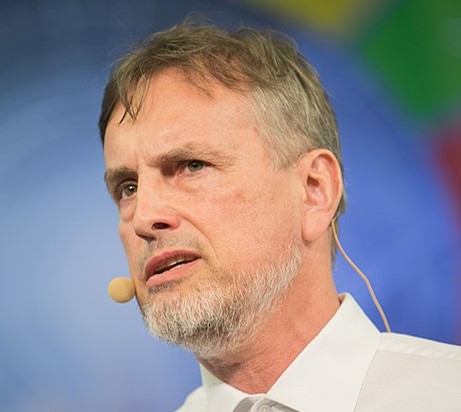} &
        \includegraphics[width=0.117\textwidth,height=0.117\textwidth]{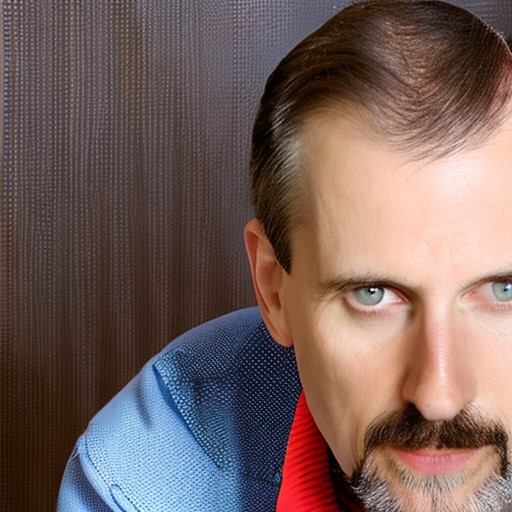} &
        \includegraphics[width=0.117\textwidth,height=0.117\textwidth]{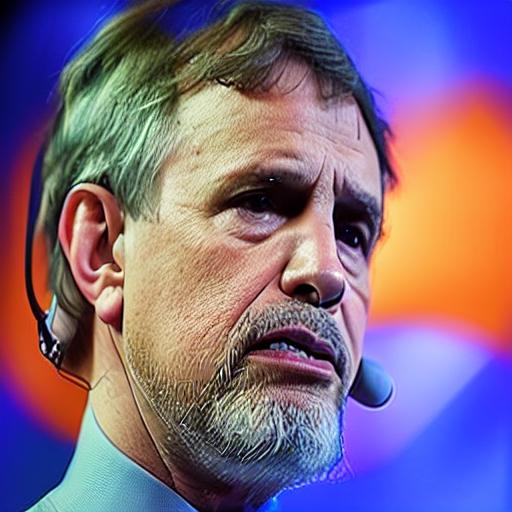} & %
        \includegraphics[width=0.117\textwidth,height=0.117\textwidth]{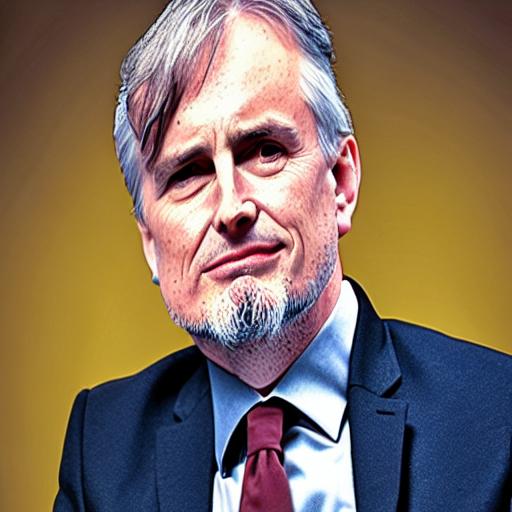} & %
        \includegraphics[width=0.117\textwidth,height=0.117\textwidth]{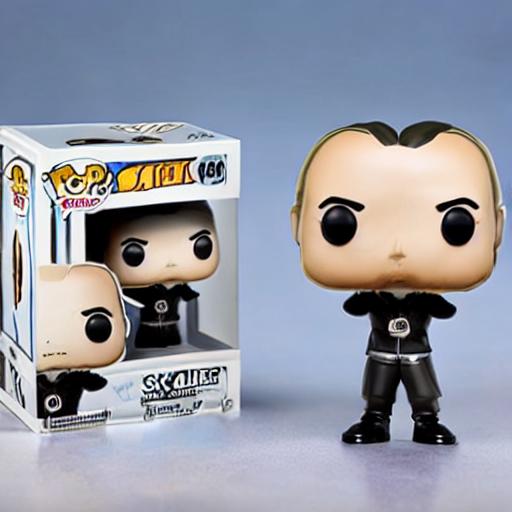} & %
        \includegraphics[width=0.117\textwidth,height=0.117\textwidth]{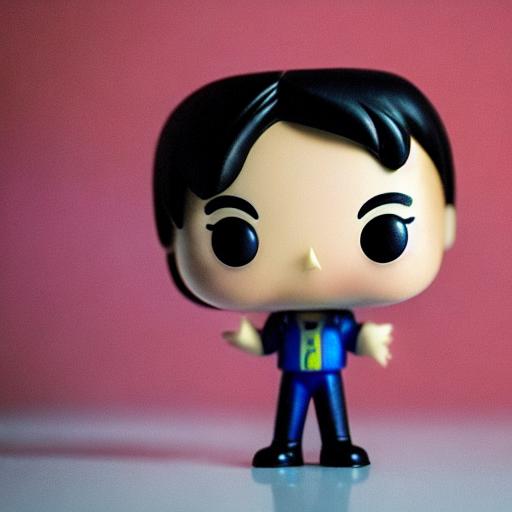}& %
        \includegraphics[width=0.117\textwidth,height=0.117\textwidth]{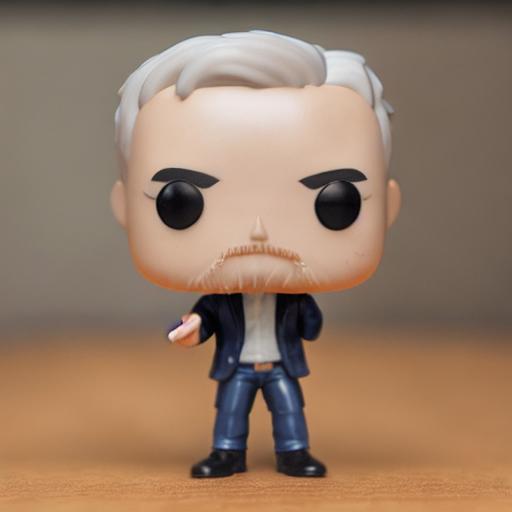} &
        \includegraphics[width=0.117\textwidth,height=0.117\textwidth]{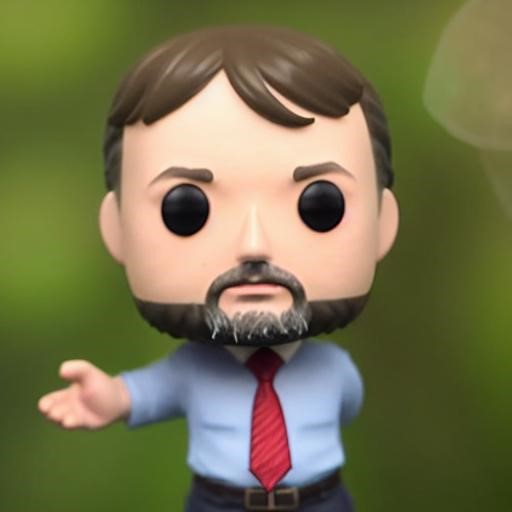} & \raisebox{0.048\textwidth}{\rotatebox[origin=t]{-90}{\scalebox{0.9}{\begin{tabular}{c@{}c@{}c@{}} \pholdercolor{} Funko Pop\end{tabular}}}} \\

        \includegraphics[width=0.117\textwidth,height=0.117\textwidth]{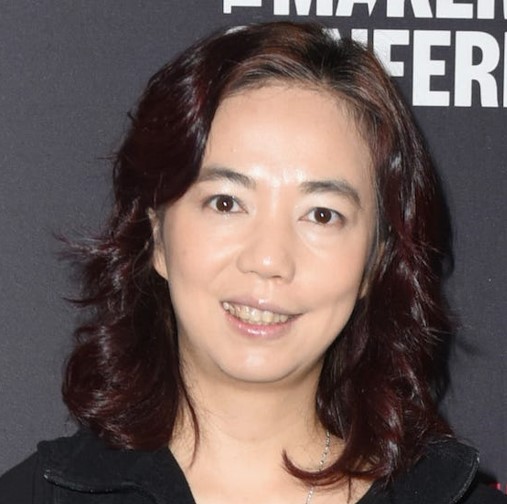} &
        \includegraphics[width=0.117\textwidth,height=0.117\textwidth]{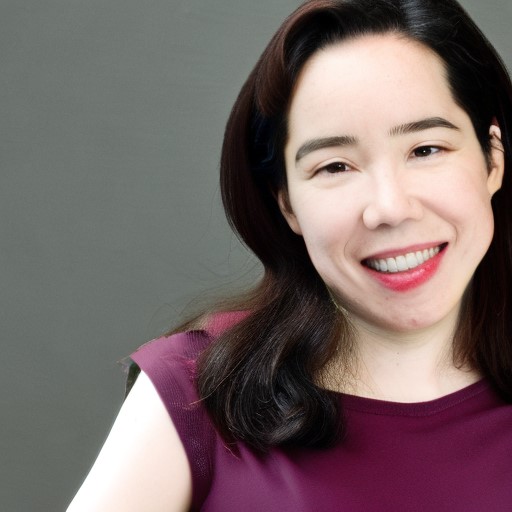} &
        \includegraphics[width=0.117\textwidth,height=0.117\textwidth]{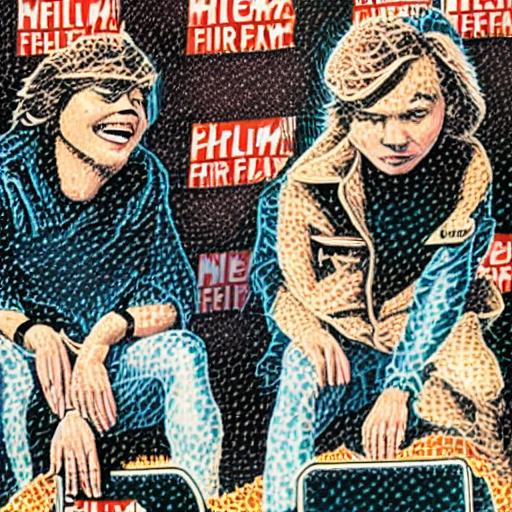} & %
        \includegraphics[width=0.117\textwidth,height=0.117\textwidth]{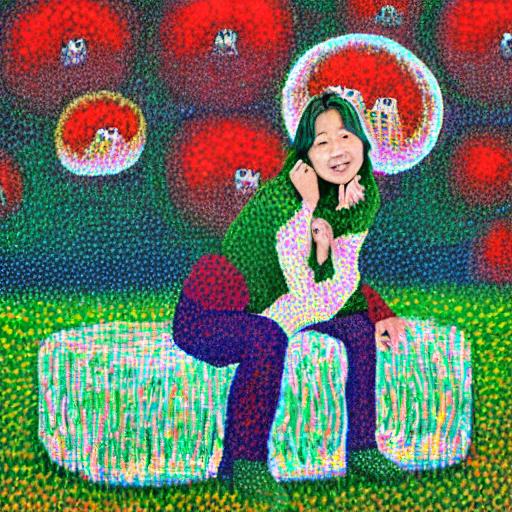} & %
        \includegraphics[width=0.117\textwidth,height=0.117\textwidth]{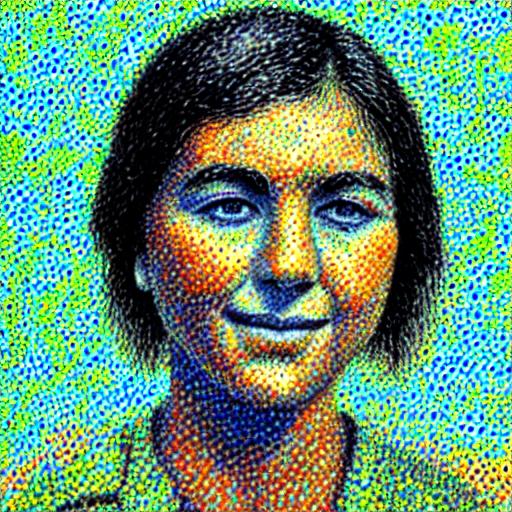} &  %
        \includegraphics[width=0.117\textwidth,height=0.117\textwidth]{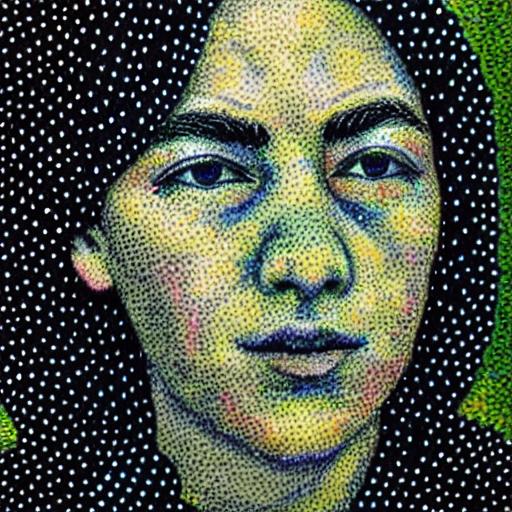}& %
        \includegraphics[width=0.117\textwidth,height=0.117\textwidth]{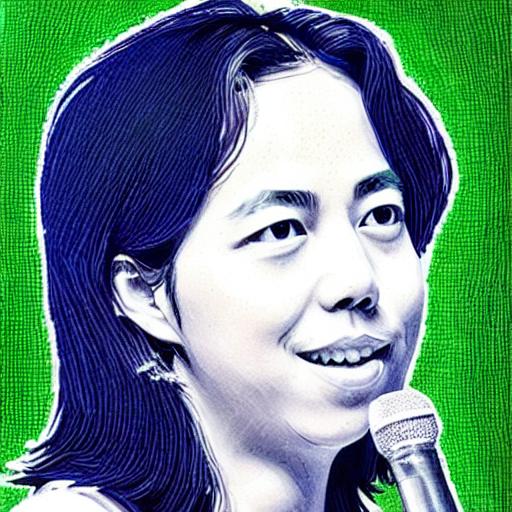} &
        \includegraphics[width=0.117\textwidth,height=0.117\textwidth]{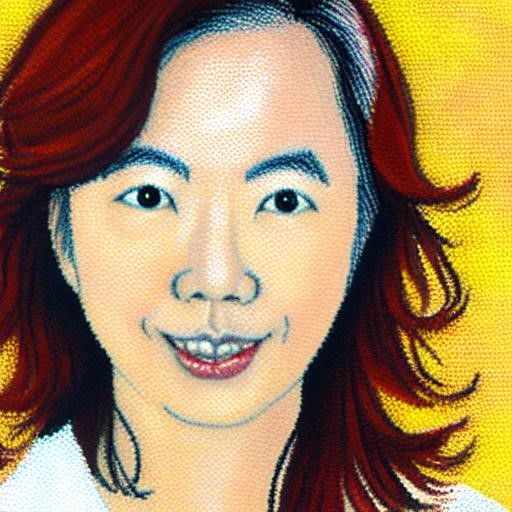} & \raisebox{0.035\textwidth}{\rotatebox[origin=t]{-90}{\scalebox{0.9}{\begin{tabular}{c@{}c@{}c@{}} Pointillism \\ painting of \pholdercolor{} \end{tabular}}}} \\

        \includegraphics[width=0.117\textwidth,height=0.117\textwidth]{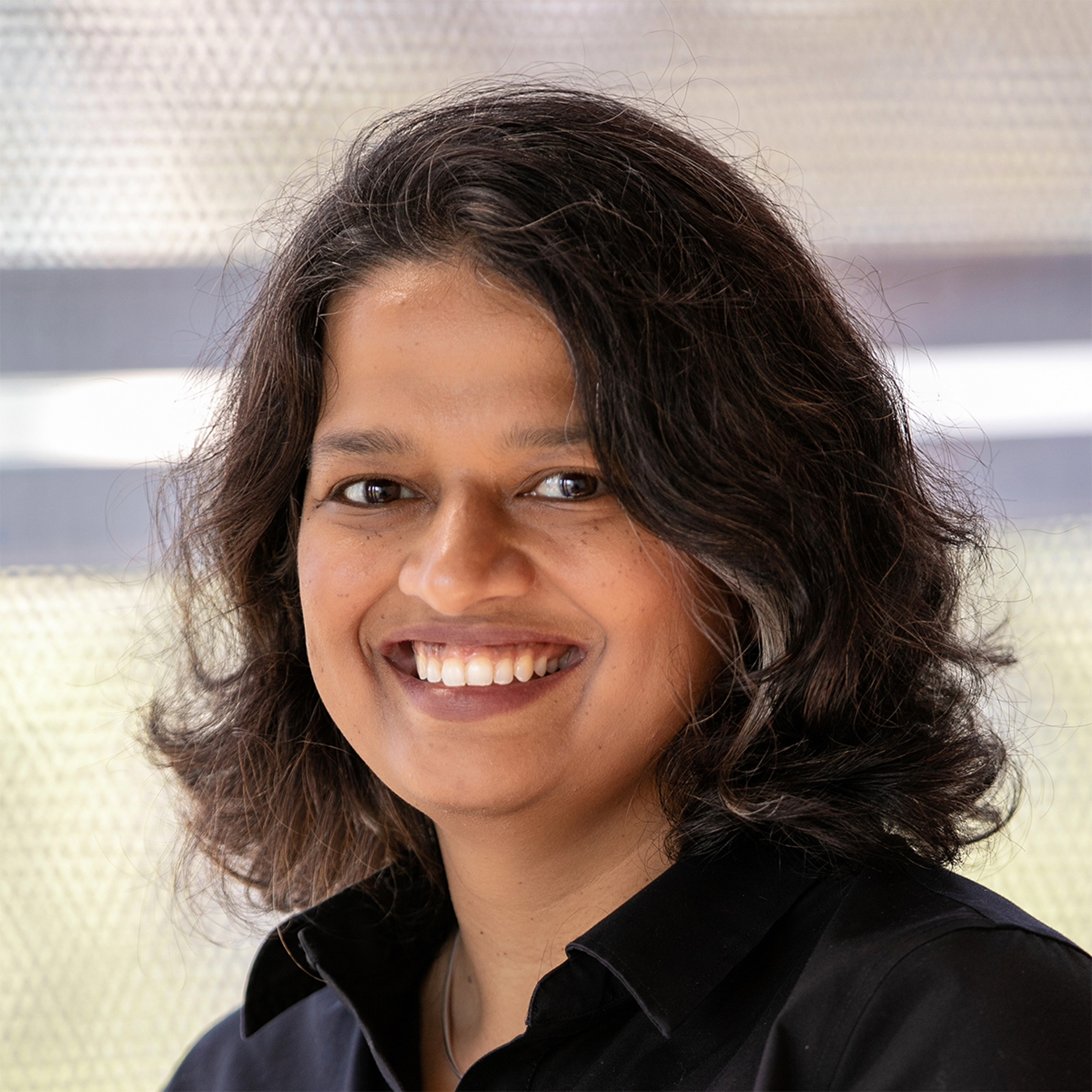} &
        \includegraphics[width=0.117\textwidth,height=0.117\textwidth]{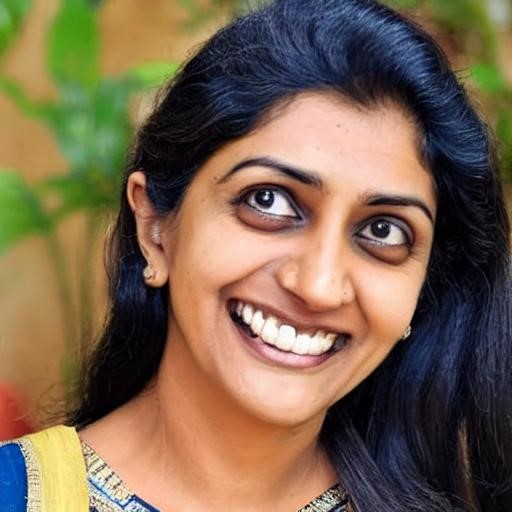} &
        \includegraphics[width=0.117\textwidth,height=0.117\textwidth]{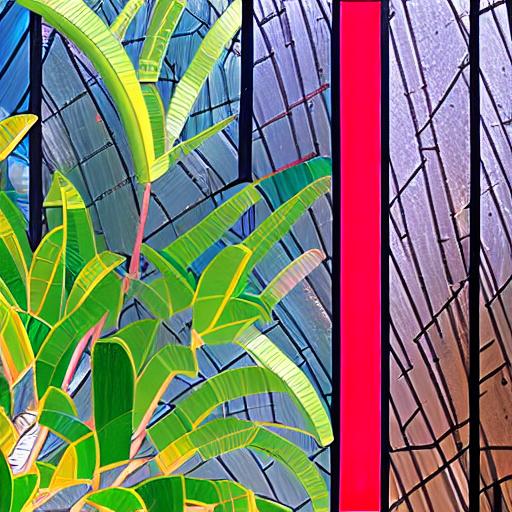} & %
        \includegraphics[width=0.117\textwidth,height=0.117\textwidth]{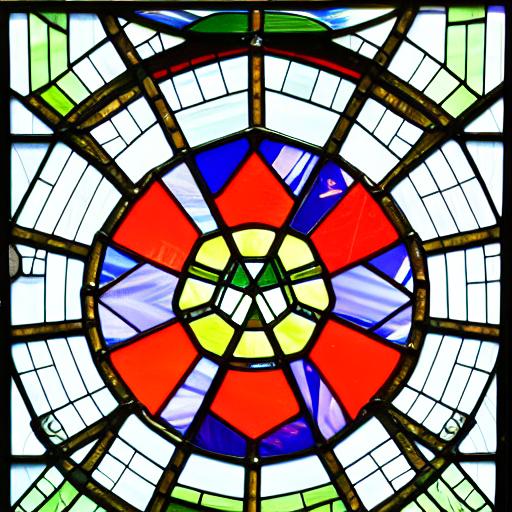} & %
        \includegraphics[width=0.117\textwidth,height=0.117\textwidth]{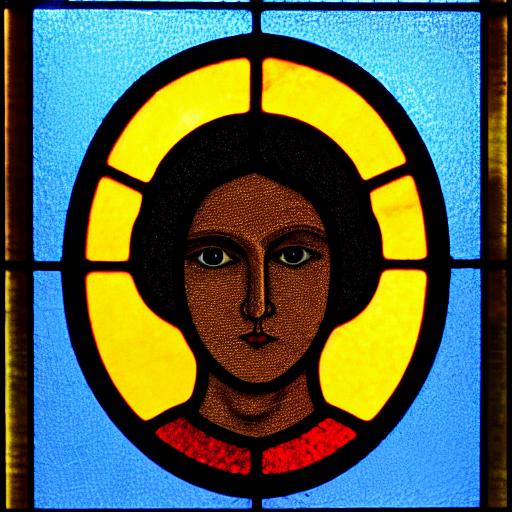} &  %
        \includegraphics[width=0.117\textwidth,height=0.117\textwidth]{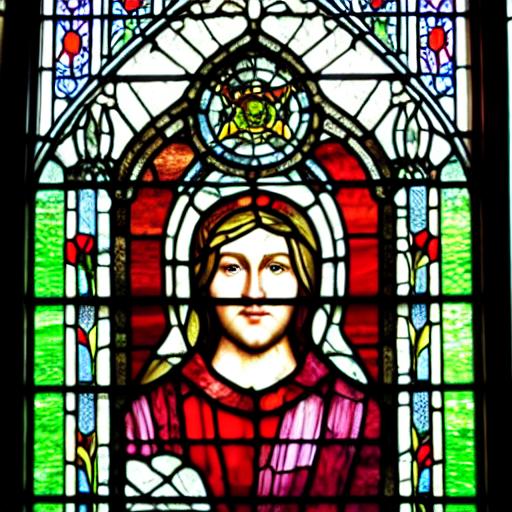}& %
        \includegraphics[width=0.117\textwidth,height=0.117\textwidth]{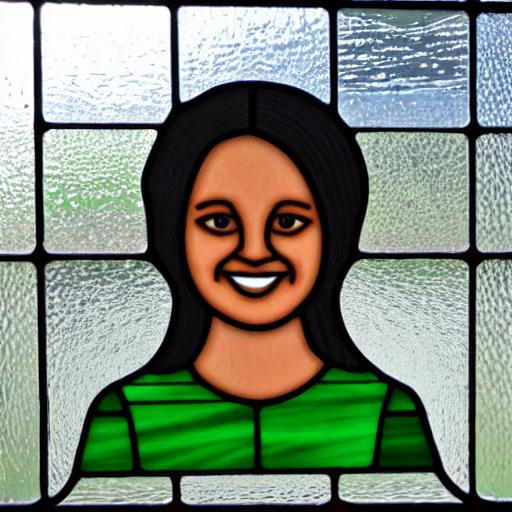} &
        \includegraphics[width=0.117\textwidth,height=0.117\textwidth]{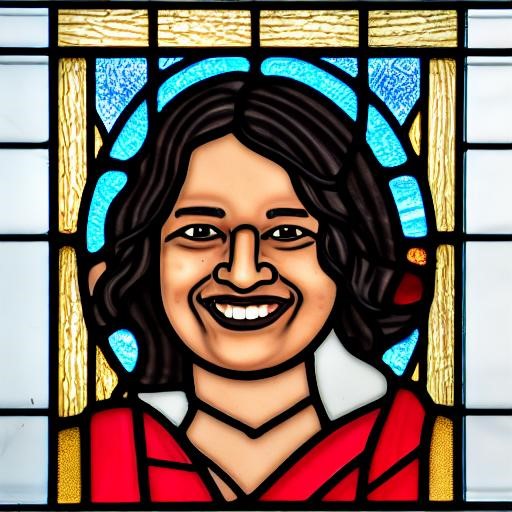} & \raisebox{0.035\textwidth}{\rotatebox[origin=t]{-90}{\scalebox{0.9}{\begin{tabular}{c@{}c@{}c@{}} \pholdercolor{} stained \\ glass window \end{tabular}}}} \\

        \includegraphics[width=0.117\textwidth,height=0.117\textwidth]{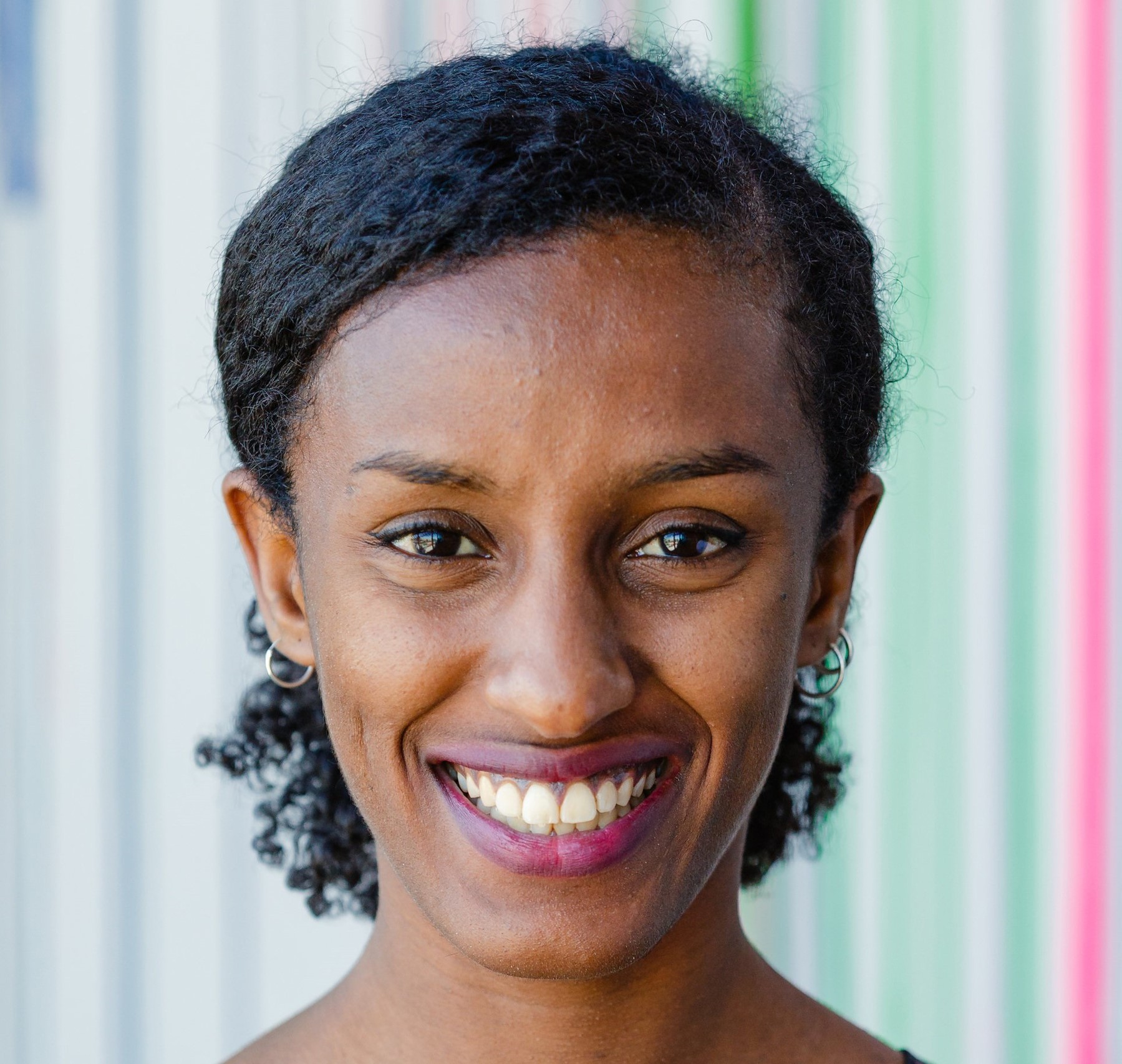} &
        \includegraphics[width=0.117\textwidth,height=0.117\textwidth]{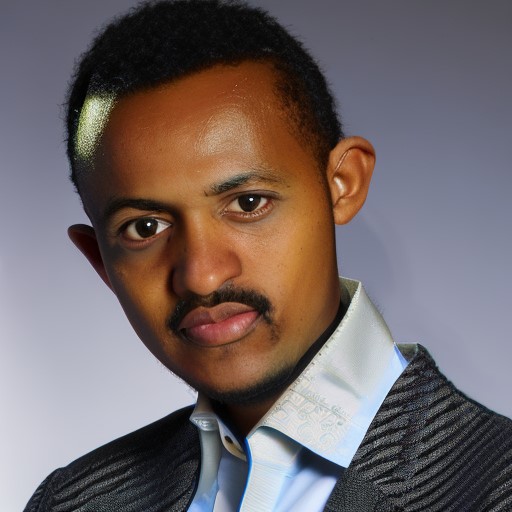} &
        \includegraphics[width=0.117\textwidth,height=0.117\textwidth]{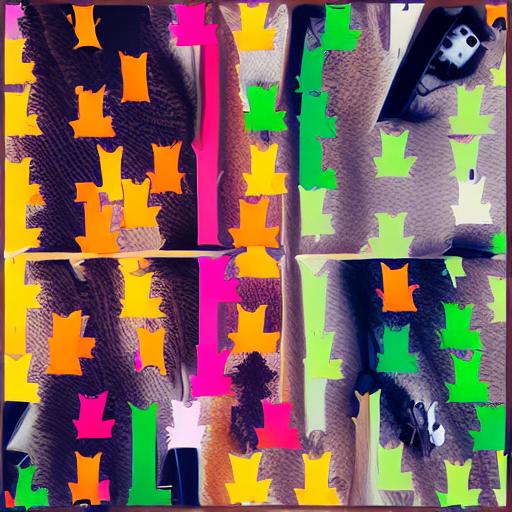} &  %
        \includegraphics[width=0.117\textwidth,height=0.117\textwidth]{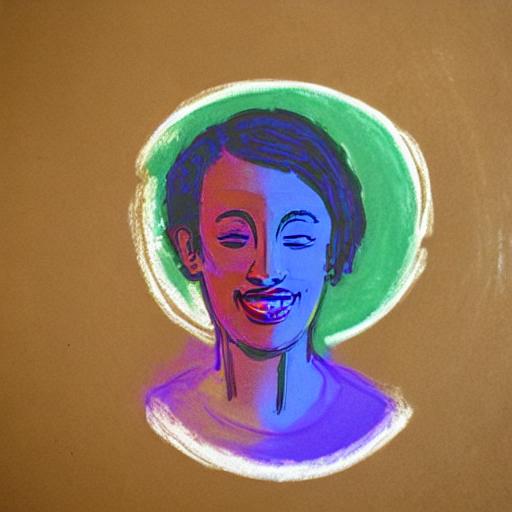} & %
        \includegraphics[width=0.117\textwidth,height=0.117\textwidth]{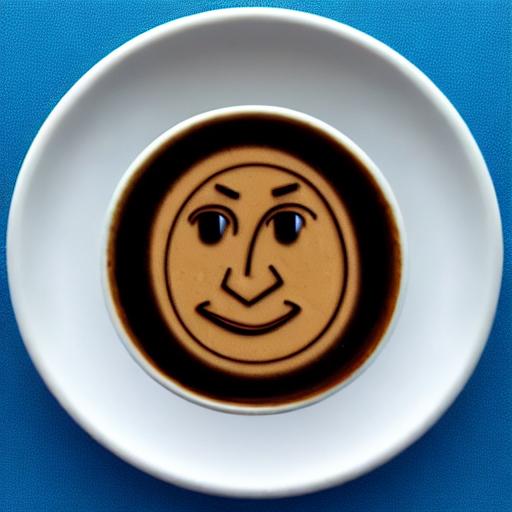} &  %
        \includegraphics[width=0.117\textwidth,height=0.117\textwidth]{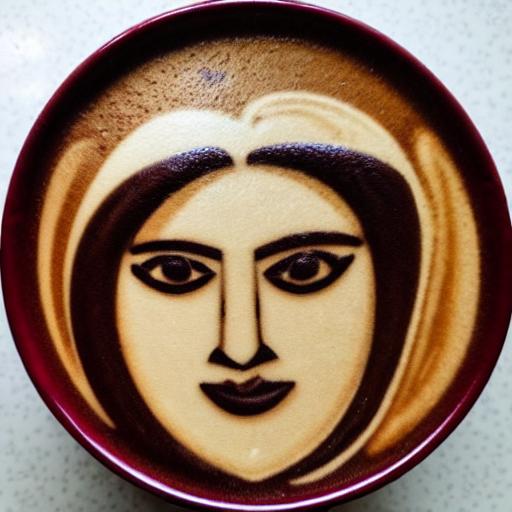}& %
        \includegraphics[width=0.117\textwidth,height=0.117\textwidth]{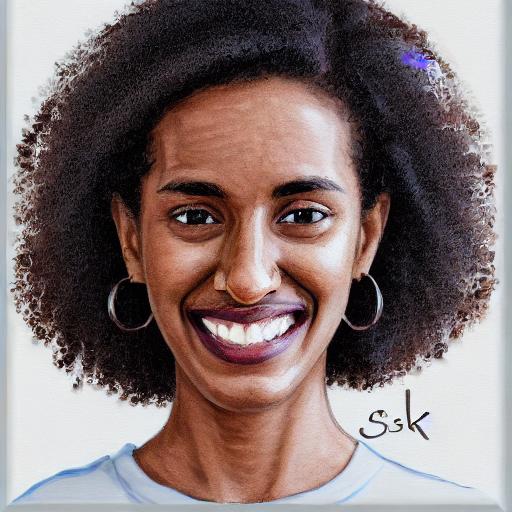} &
        \includegraphics[width=0.117\textwidth,height=0.117\textwidth]{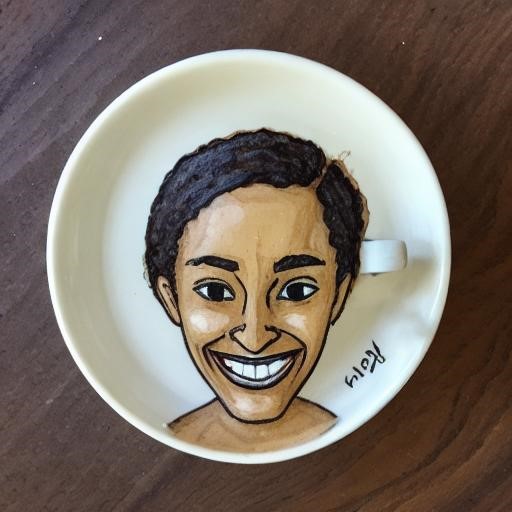} & \raisebox{0.048\textwidth}{\rotatebox[origin=t]{-90}{\scalebox{0.9}{\begin{tabular}{c@{}c@{}c@{}} \pholdercolor{} latte art \end{tabular}}}} \\

    \end{tabular}
    
    }
    \vspace{-0.2cm}
    \caption{Comparisons with prior personalization methods. We show (left to right): The image used for 1-shot personalizations, the result of prompting Stable Diffusion with the researcher's name, the results of personalization with Textual Inversion and DreamBooth using $1$- and $5$-images respectively, the results of training a DreamBooth model and an embedding concurrently on $5$ images, and finally our own result, and the driving prompt. Our method achieves comparable or better quality with only a single image and a fraction of the time.}
    \vspace{-0.25cm}
    \label{fig:qual_comp}
\end{figure*}

\section{Experiments}\label{sec:experiments}
To demonstrate the effectiveness of our approach, which we dub Encoder for Tuning (E4T), we conduct a set of comparisons against the two prior personalization methods: Textual Inversion (TI)~\citep{gal2022image} and DreamBooth (DB)~\citep{ruiz2022dreambooth}. For TI, we used the implementation provided by the authors. For DB, official code is not available. We show results using an implementation that follows the paper and tunes only the denoiser's U-Net~\citep{huggingface2022dreambooth}, as well as the results of an implementation that also tunes the word embeddings (\ie performs both DB and TI concurrently).

\paragraph{\textbf{Qualitative Evaluation}}
We begin with a qualitative evaluation, demonstrating that our method can capture a high level of detail using only a single image and a fraction of the training steps.

\Cref{fig:qual_comp} shows the results of face-personalization using the three approaches, across different levels of supervision and for a range of prompts. We follow TI and use the symbol \pholdercolor{} to represent the personalized concept in the prompts. The results of E4T are competitive or better than both baseline approaches, even when these methods have access to additional data. Note that despite training only on aligned face data, our method still enables generation of unaligned or full-body images. In \cref{fig:domains} we demonstrate that E4T can be applied to additional domains, including abstract concept classes such as artistic styles.

\paragraph{\textbf{Quantitative Evaluation}}
We evaluate our approach quantitatively using a large-scale identity preservation experiment, as typical in GAN inversion works. Here, we use our encoder and the two baselines to personalize a model for individuals taken from the LFW~\citep{huang2008labeled} dataset. We train a model on each of the test-set identities that contain between 3 and 10 images. Our model uses only a single randomly chosen image for each identity. Competing methods use either the same single image (``single") or the entire 3-10 image set (``multi").
In total, we train a total of 232 models for each baseline and level of supervision.

\begin{figure*}[!hbt]
    \setlength{\abovecaptionskip}{7.5pt}
    \setlength{\belowcaptionskip}{-0.5pt}
    \setlength{\tabcolsep}{0.55pt}
    \centering
\begin{tabular}{c c}

\includegraphics[width=0.47\textwidth]{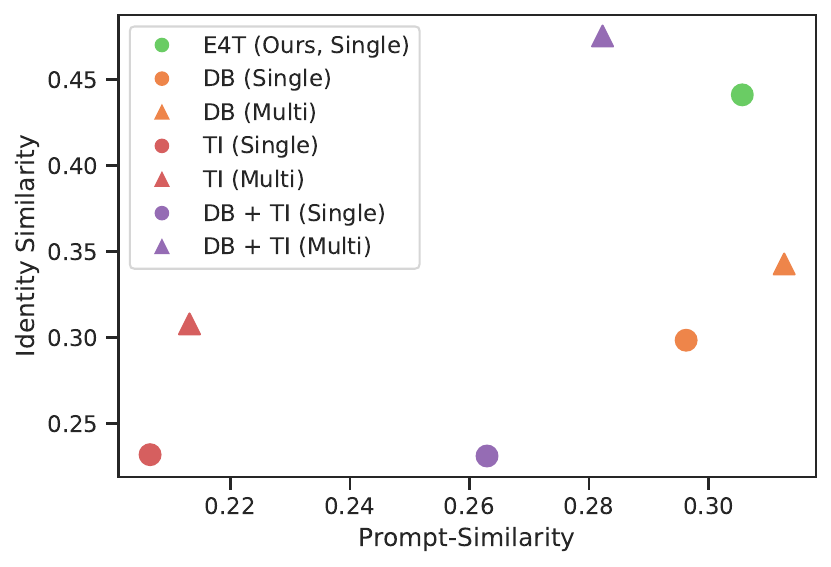}
 &
\includegraphics[width=0.47\textwidth]{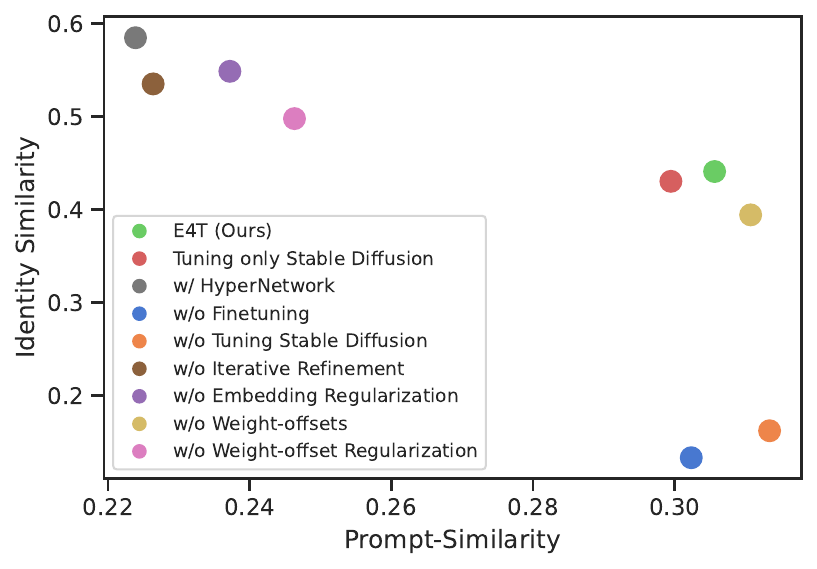} \\[1pt]
\quad\quad\quad(a) & \quad\quad\quad(b) 
\end{tabular}

\caption{(a) We compare our method to existing personalization approaches. Our method (green) significantly outperforms other methods with a single image, and sits on the pareto front when compared to few-shot personalization approaches, despite being significantly quicker. (b) Ablation study results. Our full method (green) achieves high identity preservation while maintaining editability.}
\label{fig:quantitative}
\end{figure*}

We then generate a set of images of every identity across a range of prompts, covering a range of modifications such as full-body shots, stylization, accessorizing and background changes. We measure identity preservation by computing the average pair-wise identity similarity~\citep{huang2020curricularface} between each person's training set and the generated results. Following~\citep{gal2022image}, we further measure prompt-adherance by computing the average CLIP-space similarity~\citep{radford2021learning} between each generated image and its concept-less prompt. The results are shown in \cref{fig:quantitative}(a). Our method sits on an appealing point on the pareto-front, representing both high identity-preservation and prompt-adherence, demonstrating that it can be used to effectively capture identities at a fraction of the time. 
\begin{wraptable}[7]{R}{0.5\linewidth}
\vspace{-15pt}
\setlength{\tabcolsep}{3pt}
\setlength{\abovecaptionskip}{1.5pt}
\setlength{\belowcaptionskip}{-4.5pt}
\footnotesize
\caption{Training iterations and times for personalization methods. \\
$^*$not including setup time.}\label{tab:personalization_times}
\begin{tabular}{l c c}
Method & Iterations & Time (seconds)$^*$ \\
\midrule 
TI & $5,000$ & $1,517$ \\
DB & $800-1,200$ & $601-897$ \\
Ours & $\mathbf{5}$-$\mathbf{15}$ & $\mathbf{11}$-$\mathbf{30}$ \\
\end{tabular}
\end{wraptable}

In \cref{tab:personalization_times} we report the average personalization times using each method. Note that DB and E4T require different numbers of iterations for different domains. E4T is significantly quicker than the alternatives.

\begin{figure}[!hbt]
    \centering
    \setlength{\abovecaptionskip}{9.5pt}
    \setlength{\belowcaptionskip}{-3.5pt}
    \setlength{\tabcolsep}{0.55pt}
    \renewcommand{\arraystretch}{1.0}
    {\scriptsize
    \begin{tabular}{c@{\hskip 2pt} c@{\hskip 2pt} c c c}
        \begin{tabular}{c}
        \includegraphics[width=0.110\textwidth,height=0.110\textwidth]{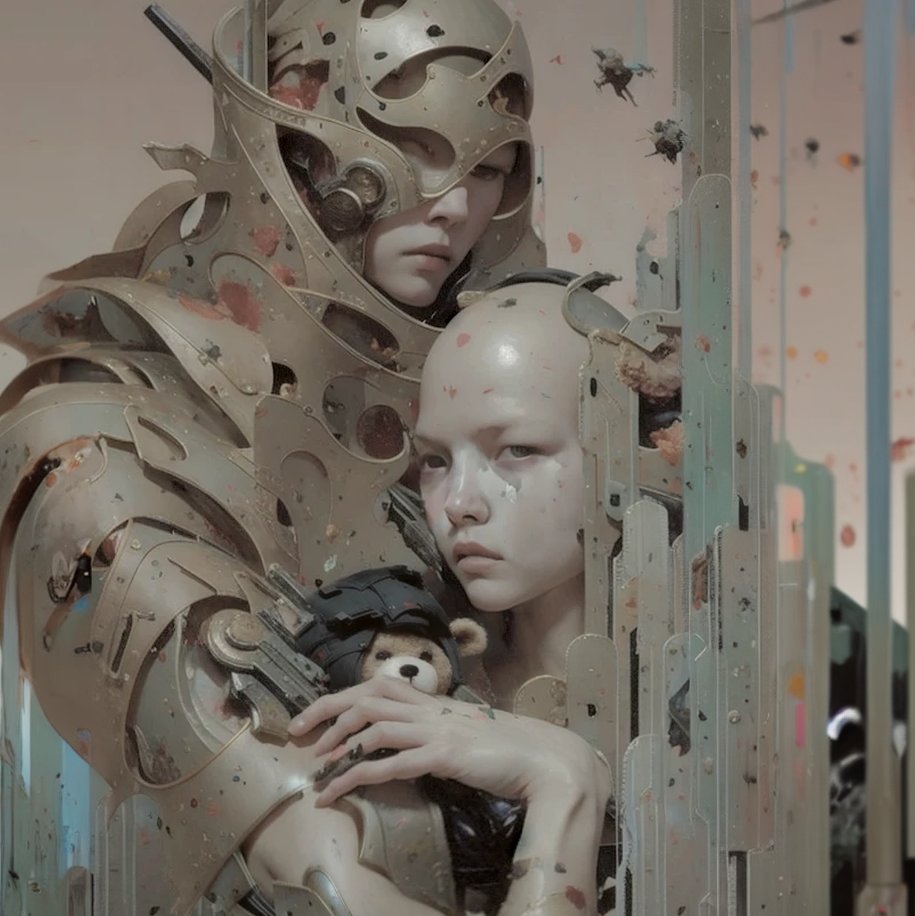} 
        \end{tabular}
        
        &
        $\rightarrow$
        &
        \begin{tabular}{c}
        \includegraphics[width=0.110\textwidth,height=0.110\textwidth]{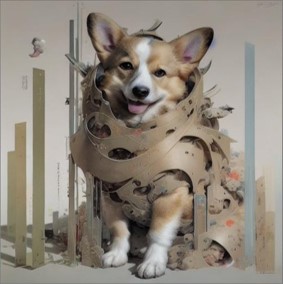} 
        \end{tabular} &
        \begin{tabular}{c}
        \includegraphics[width=0.110\textwidth,height=0.110\textwidth]{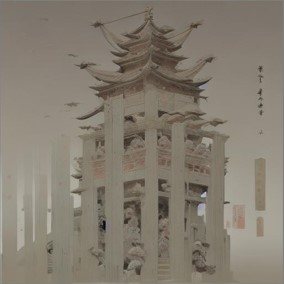}
        \end{tabular} & 
        \begin{tabular}{c}
        \includegraphics[width=0.110\textwidth,height=0.110\textwidth]{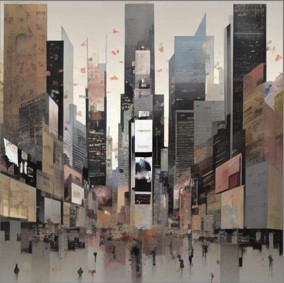}
        \end{tabular} \\
        
        \begin{tabular}{c}
        \includegraphics[width=0.110\textwidth,height=0.110\textwidth]{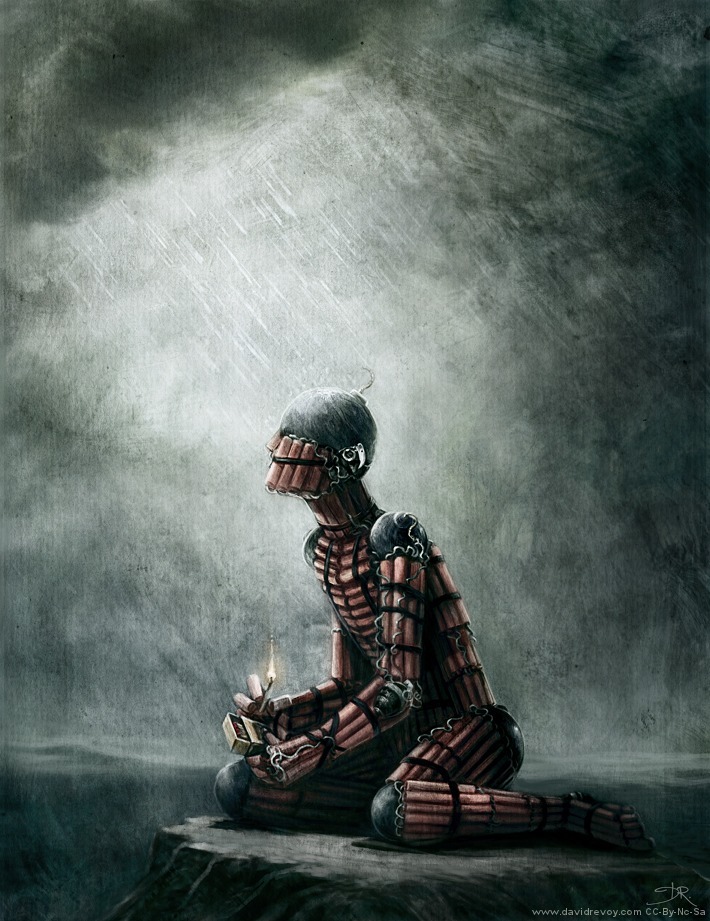} 
        \end{tabular}
        
        &
        $\rightarrow$
        &
        \begin{tabular}{c}
        \includegraphics[width=0.110\textwidth,height=0.110\textwidth]{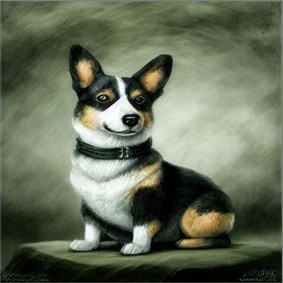} 
        \end{tabular} &
        \begin{tabular}{c}
        \includegraphics[width=0.110\textwidth,height=0.110\textwidth]{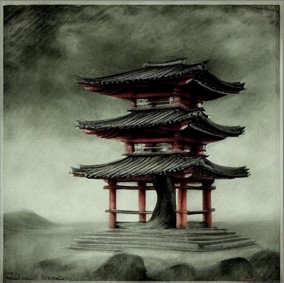} 
        \end{tabular} &
        \begin{tabular}{c}
        \includegraphics[width=0.110\textwidth,height=0.110\textwidth]{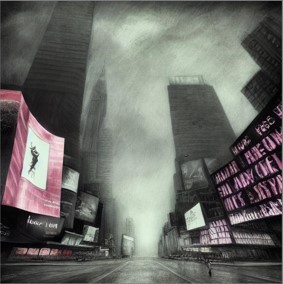}
        \end{tabular} \\
        
        {Input image} & & {\begin{tabular}{c@{}c@{}c@{}c@{}} ``Painting of an adorable \\ corgi in the style of \pholdercolor" \end{tabular}} & {\begin{tabular}{c@{}c@{}c@{}c@{}} ``Edo period pagoda \\ in the style of \pholdercolor" \end{tabular}} & {\begin{tabular}{c@{}c@{}c@{}c@{}} ``Times square \\ in the style of \pholdercolor" \end{tabular}} 
        \end{tabular} \\[3pt]

    \begin{tabular}{c@{\hskip 2pt} c@{\hskip 2pt} c c c}
        \begin{tabular}{c}
        \includegraphics[width=0.110\textwidth,height=0.110\textwidth]{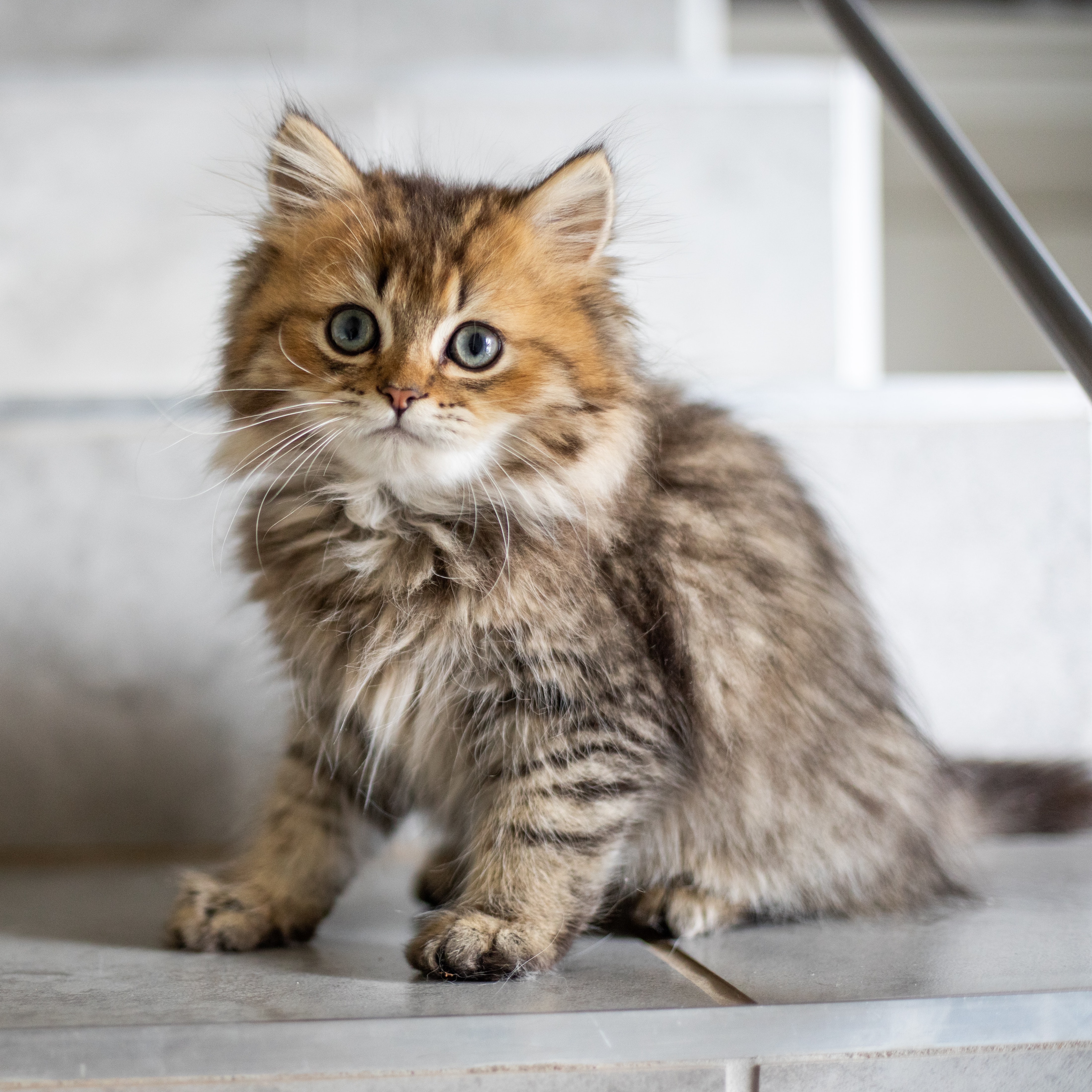} 
        \end{tabular}
        
        &
        $\rightarrow$
        &
        \begin{tabular}{c}
        \includegraphics[width=0.110\textwidth,height=0.110\textwidth]{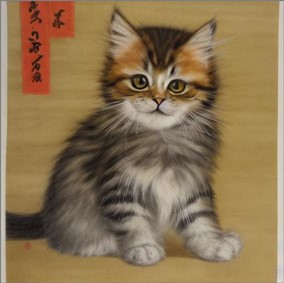} 
        \end{tabular} &
        \begin{tabular}{c}
        \includegraphics[width=0.110\textwidth,height=0.110\textwidth]{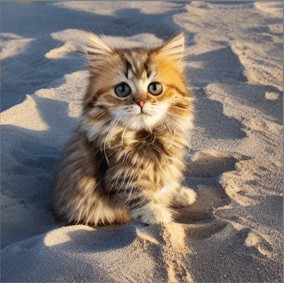} 
        \end{tabular} &
        \begin{tabular}{c}
        \includegraphics[width=0.110\textwidth,height=0.110\textwidth]{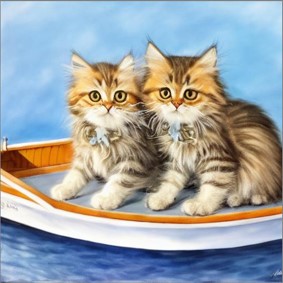}
        \end{tabular}  \\
        
        \begin{tabular}{c}
        \includegraphics[width=0.110\textwidth,height=0.110\textwidth]{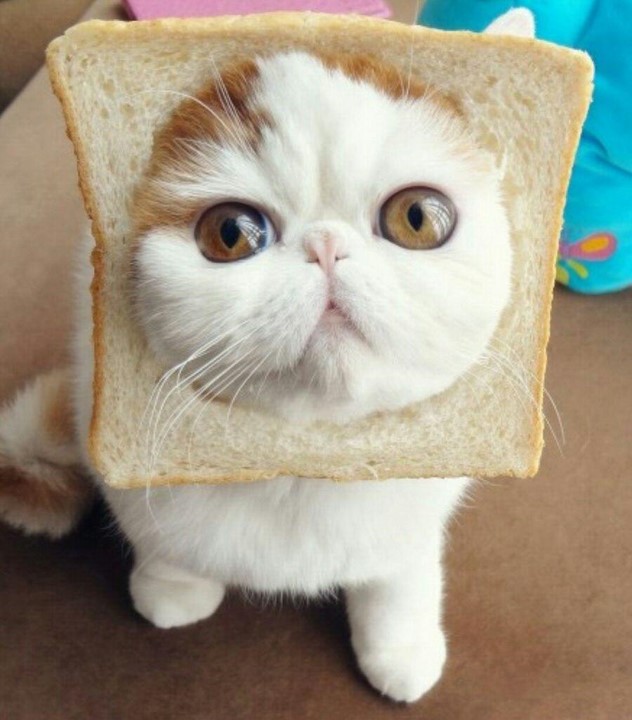} 
        \end{tabular}
        
        &
        $\rightarrow$
        &
        \begin{tabular}{c}
        \includegraphics[width=0.110\textwidth,height=0.110\textwidth]{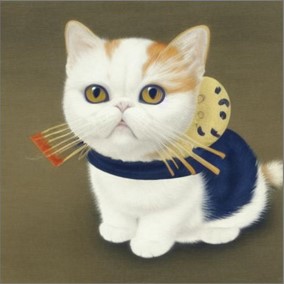}
        \end{tabular} &
        \begin{tabular}{c}
        \includegraphics[width=0.110\textwidth,height=0.110\textwidth]{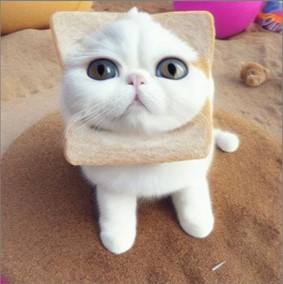} 
        \end{tabular} &
        \begin{tabular}{c}
        \includegraphics[width=0.110\textwidth,height=0.110\textwidth]{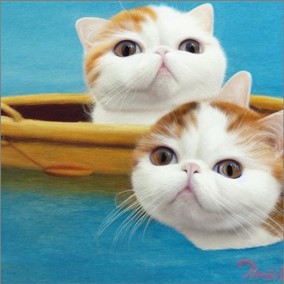} 
        \end{tabular}  \\
        
        {Input image} & & {\begin{tabular}{c@{}c@{}c@{}c@{}} ``Ukiyo-e painting \\ of \pholdercolor" \end{tabular}} & {\begin{tabular}{c@{}c@{}c@{}c@{}} ``A photo of \pholdercolor \\ on the beach" \end{tabular}} & {\begin{tabular}{c@{}c@{}c@{}c@{}} ``A painting of two \\ \pholdercolor{} on a boat" \end{tabular}} \\
    \end{tabular}}
    \caption{Our method can be applied to additional domains, including those that represent abstract concepts like artistic styles. Image credits: \href{https://twitter.com/amli_art}{@AmliArt}, \href{https://commons.wikimedia.org/wiki/User:Deevad}{@David Revoy}, \href{https://unsplash.com/photos/bhonzdJMVjY}{@Jeanie}.}
    \label{fig:domains} 
\end{figure}

\paragraph{\textbf{Ablation study}}
Our approach required a significant number of design choices. Here, we study these choices and demonstrate their importance in achieving high-fidelity results.

We begin with a qualitative evaluation using the same image-set and metrics of \cref{fig:quantitative}(a). We examine the effects of the following changes: (1) Removing the fine-tuning step, (2) tuning only our components (encoder, offsets) or only the denoiser, (2) removing the iterative refinement module, (3) removing the embedding regularization loss, (4) learning weight-offsets directly (rather than through a regularized network), and (5) training only an encoder (with no weight-offsets). We further compare to a HyperNetwork~\cite{ha2016hypernetworks} baseline, where the constant-offsets are replaced by the predictions of a HyperNetwork that feeds the aggregated CLIP-backbone features into our weight-offset prediction architecture.

The results are shown in \cref{fig:quantitative}(b). As can be observed, our full model enables both high-concept similarity and good editability. Removing the embedding regularization leads to predicted codes that reside in difficult-to-modify regions of the latent space, esentially overfitting the personalized model to the given concept. A similar effect can be observed when removing the regularization on the weight-offsets, or when using a HyperNetwork. Here, the extra degrees of freedom lead to quickly overfitting on the target domain. Avoiding the inference-time tuning step or restricting it only to our components leads to significant reduction in identity preservation. Finally, discarding the iterative-refinement module harms the model's ability to modify the concept, highlighting the advantage of being able to focus on different aspects of the target at each synthesis step. We note that similar observations were made by E-Diffi~\cite{balaji2022ediffi} which demonstrated that using different denoisers at different stages of the cleaning process can help improve visual fidelity and prompt-matching. 

\paragraph{\textbf{Refinement analysis}} 
Our iterative refinement approach provides us with a window into the denoising process, allowing us to see which regions of the image the network considers most worth-while to focus on at each denoising stage.

\begin{figure*}[!htbp]
 \setlength{\abovecaptionskip}{6.5pt}
 \setlength{\belowcaptionskip}{3.5pt}
\centering
\includegraphics[width=0.9\textwidth]{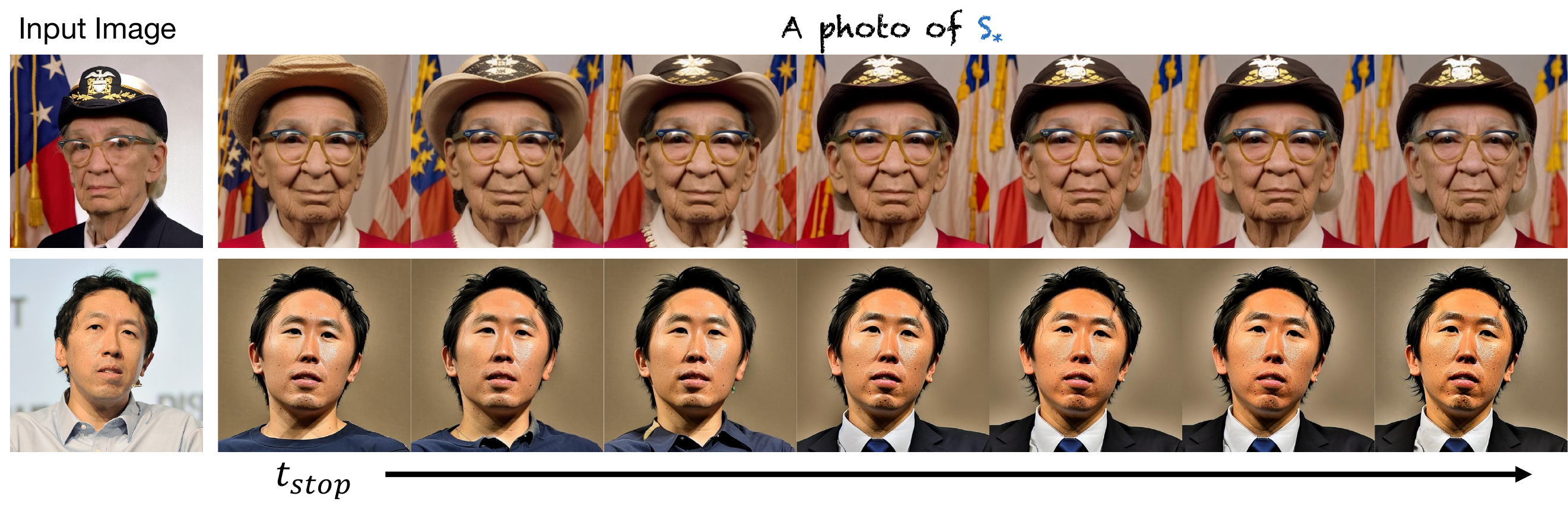}
\label{fig:refinement}
\caption{We show the effect of the iterative refinement prediction by freezing our encoder's predicted word-embedding at a specific denoising step $t_{stop}$. Early on, the model captures high-level details like the shape of the face or prominent accessories. Further steps refine the facial features and the image semantics (\eg portraying a military hat instead of just a hat).}
\end{figure*}

To do so, we employ our concept-tuned encoder and model pair to generate novel images of the concept, where we freeze the encoder's predictions once we reach an intermediate time step, $t_{stop}$. In \cref{fig:refinement} we show results along various values of $t_{stop}$. As can be seen, in the early steps, the network prefers to focus on high-level semantics, such as the shape of the head or color schemes. As the process continues, we observe that the network shifts its focus to finer details like the layout of the hair.

\begin{figure}[!b]
\centering
     \setlength{\abovecaptionskip}{6.5pt}
     \setlength{\belowcaptionskip}{3.5pt}
\includegraphics[width=0.95\linewidth]{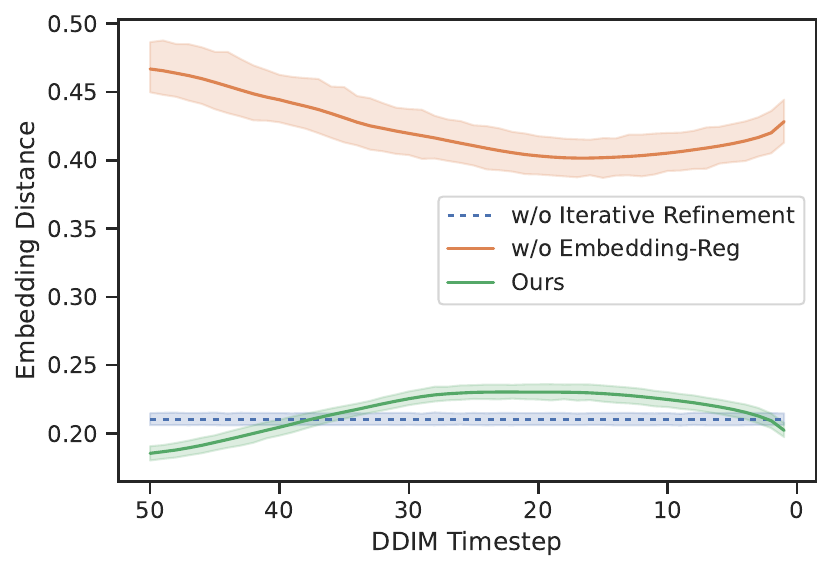}
\label{fig:embedding_change}
\caption{The distance between our predicted embeddings and the coarse class embedding, as a function of the denoising time-step. Iterative refinement enables the network to focus on high-level details in early steps, remaining closer to the domain of real-words. When unique-concept details become more relevant, the distance increases. Without regularization, embeddings deviate significantly from the real-word domain, leading to overfitting and poor editability.}
\end{figure}

We further analyze the distance between our predicted embeddings and those of the coarse class, with and without the embedding regularization. The results are provided in \cref{fig:embedding_change}. With regularization, we notice that the embeddings begin small, as there is less need to deviate from the core word distribution in order to match rough features like the shape of the head. When adding finer details, the embeddings increase in order to capture the person-specific semantics. At the final steps, the denoiser relies mostly on the image features~\citep{balaji2022ediffi} and so the embedding's role is no longer needed.
In contrast, without regularization, the encoder predicts a large embedding right away. We hypothesize that in doing so, it attempts to force a structure on the initial noise which will be cleaned into a semblance of the target (in a similar manner as DDIM inversion~\citep{song2020denoising,dhariwal2021diffusionBeatsGAN}). This gives the denoiser very little room to deviate, and leads to overfitting the concept.

\paragraph{\textbf{Limitations}}

\begin{figure}[hbp]
    \centering
    \setlength{\abovecaptionskip}{6.5pt}
    \setlength{\belowcaptionskip}{-3.5pt}
    \setlength{\tabcolsep}{0.55pt}
    \renewcommand{\arraystretch}{1.0}
    {\scriptsize
    \begin{tabular}{c}
    \begin{tabular}{c@{\hskip 2pt} c@{\hskip 2pt} c c c}
        \begin{tabular}{c}
        \includegraphics[width=0.110\textwidth,height=0.110\textwidth]{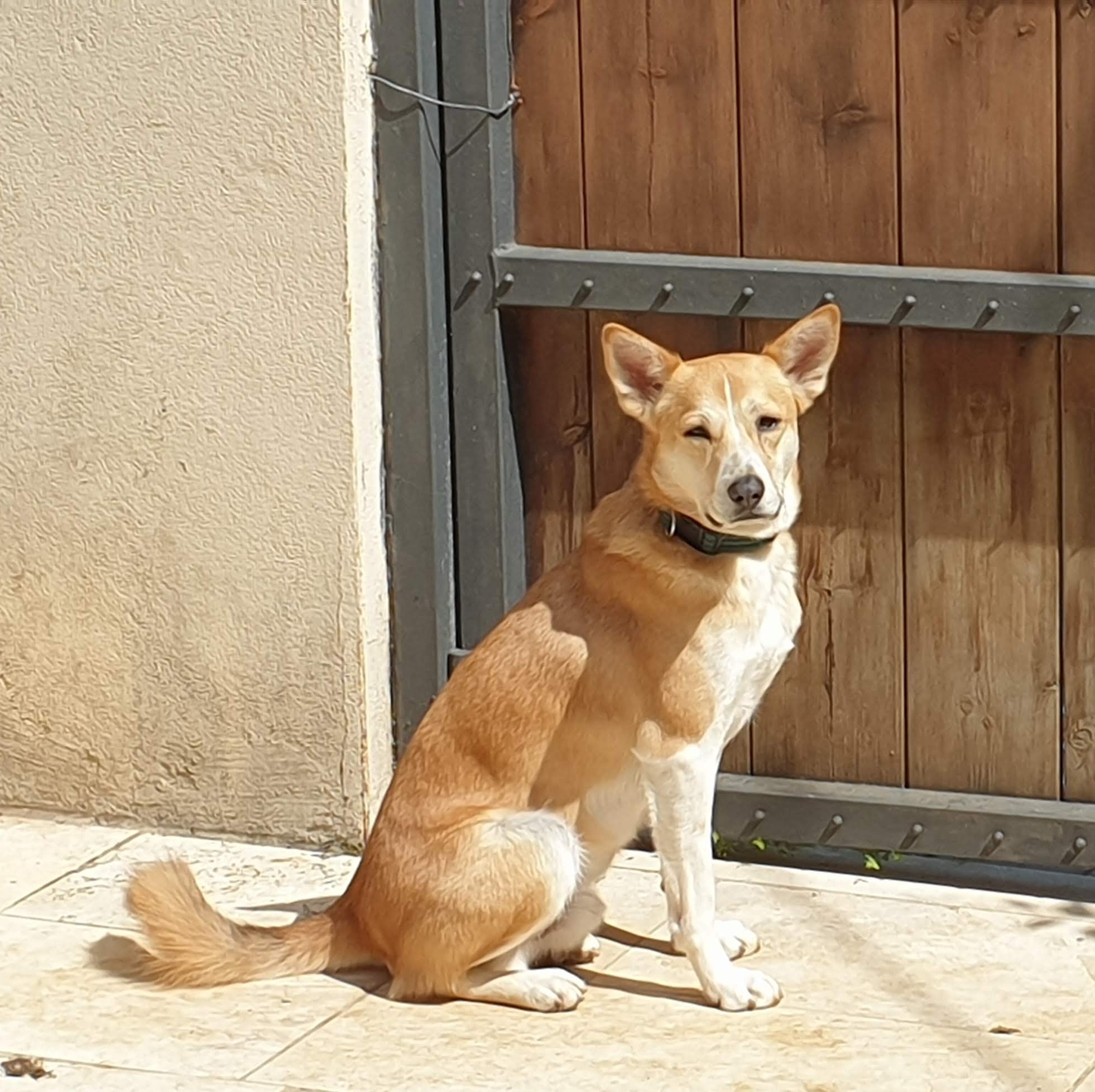} 
        \end{tabular}
        
        &
        $\rightarrow$
        &
        \begin{tabular}{c}
        \includegraphics[width=0.110\textwidth,height=0.110\textwidth]{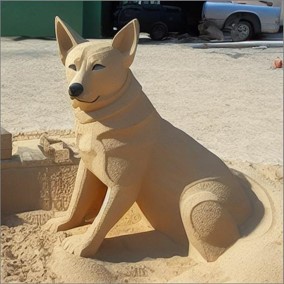} 
        \end{tabular} &
        \begin{tabular}{c}
        \includegraphics[width=0.110\textwidth,height=0.110\textwidth]{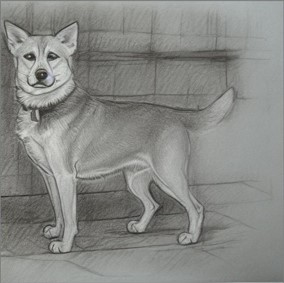}
        \end{tabular} & 
        \begin{tabular}{c}
        \includegraphics[width=0.110\textwidth,height=0.110\textwidth]{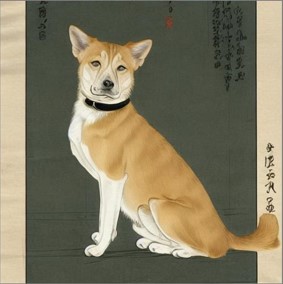}
        \end{tabular} \\
        
        \begin{tabular}{c}
        \includegraphics[width=0.110\textwidth,height=0.110\textwidth]{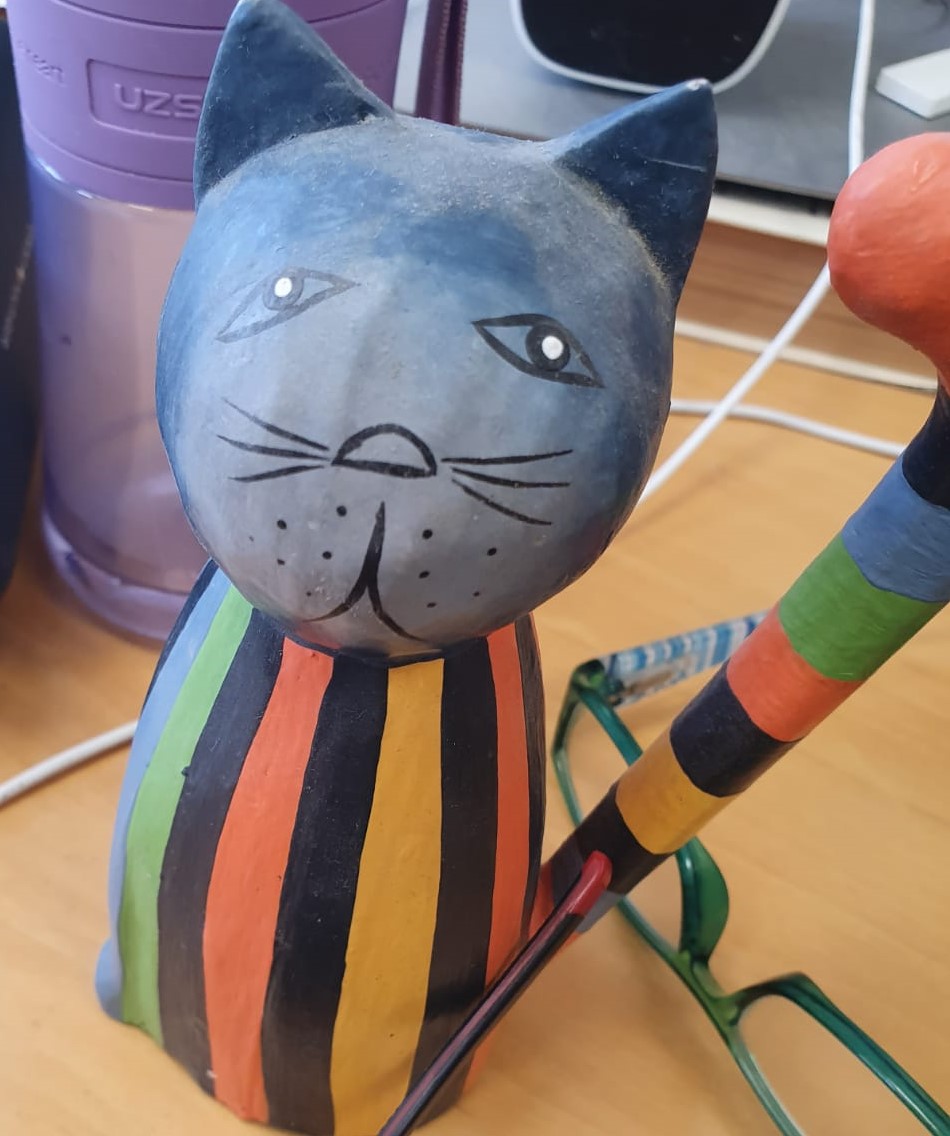} 
        \end{tabular}
        
        &
        $\rightarrow$
        &
        \begin{tabular}{c}
        \includegraphics[width=0.110\textwidth,height=0.110\textwidth]{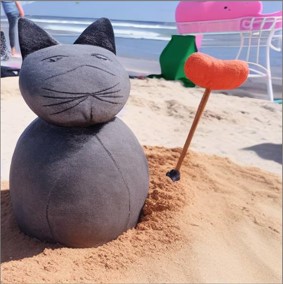} 
        \end{tabular} &
        \begin{tabular}{c}
        \includegraphics[width=0.110\textwidth,height=0.110\textwidth]{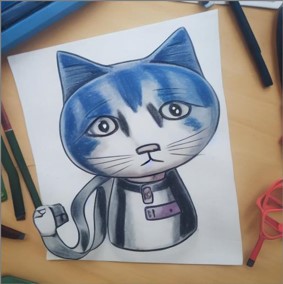} 
        \end{tabular} &
        \begin{tabular}{c}
        \includegraphics[width=0.110\textwidth,height=0.110\textwidth]{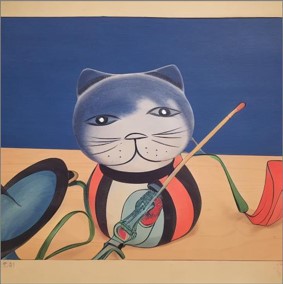}
        \end{tabular} \\
        
        {Input image} & & {\begin{tabular}{c@{}c@{}c@{}c@{}} ``\pholdercolor{} sand sculpture" \end{tabular}} & {\begin{tabular}{c@{}c@{}c@{}c@{}} ``Manga drawing \\ of \pholdercolor" \end{tabular}} & {\begin{tabular}{c@{}c@{}c@{}c@{}} ``Ukiyo-e painting \\ of \pholdercolor" \end{tabular}} 
        \end{tabular} \\ \\

    \begin{tabular}{c@{\hskip 2pt} c@{\hskip 2pt} c c c}
        \begin{tabular}{c}
        \includegraphics[width=0.110\textwidth,height=0.110\textwidth]{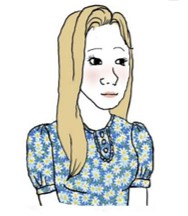} 
        \end{tabular}
        
        &
        $\rightarrow$
        &
        \begin{tabular}{c}
        \includegraphics[width=0.110\textwidth,height=0.110\textwidth]{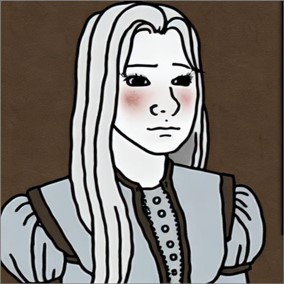} 
        \end{tabular} &
        \begin{tabular}{c}
        \includegraphics[width=0.110\textwidth,height=0.110\textwidth]{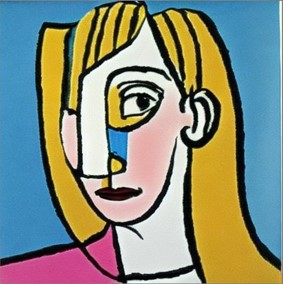} 
        \end{tabular} &
        \begin{tabular}{c}
        \includegraphics[width=0.110\textwidth,height=0.110\textwidth]{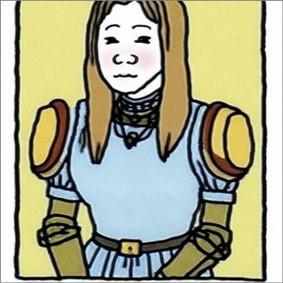}
        \end{tabular}  \\
        
        \begin{tabular}{c}
        \includegraphics[width=0.110\textwidth,height=0.110\textwidth]{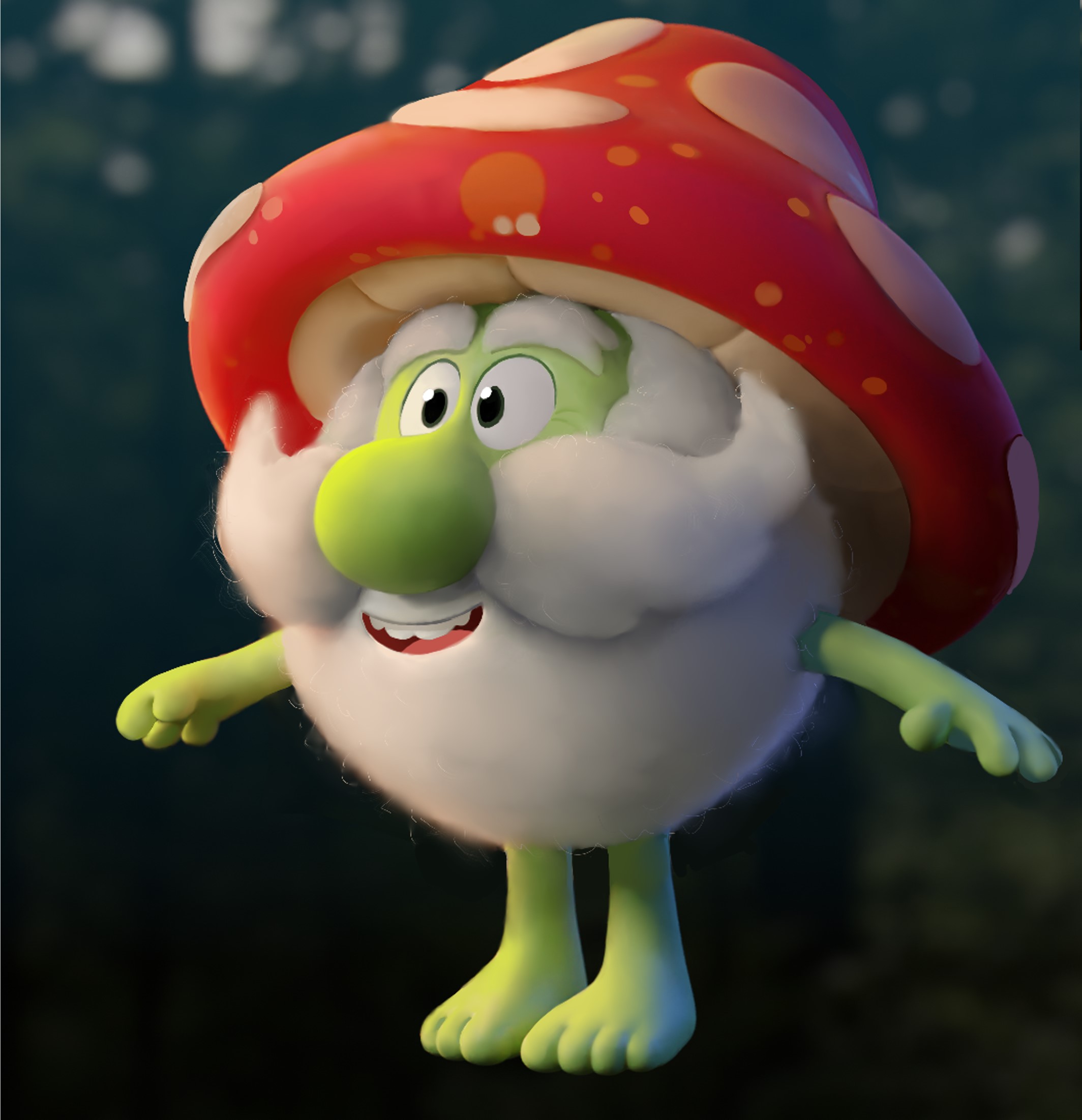} 
        \end{tabular}
        
        &
        $\rightarrow$
        &
        \begin{tabular}{c}
        \includegraphics[width=0.110\textwidth,height=0.110\textwidth]{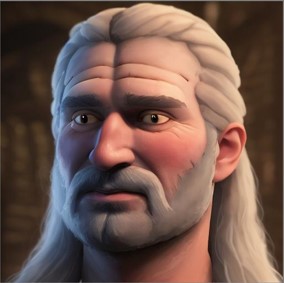}
        \end{tabular} &
        \begin{tabular}{c}
        \includegraphics[width=0.110\textwidth,height=0.110\textwidth]{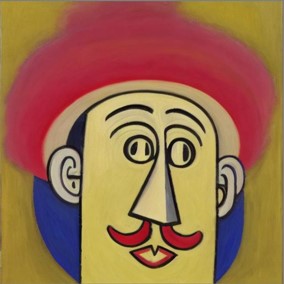} 
        \end{tabular} &
        \begin{tabular}{c}
        \includegraphics[width=0.110\textwidth,height=0.110\textwidth]{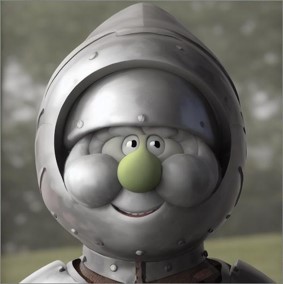} 
        \end{tabular}  \\
        
        {Input image} & & {\begin{tabular}{c@{}c@{}c@{}c@{}} ``\pholdercolor{} as a witcher" \end{tabular}} & {\begin{tabular}{c@{}c@{}c@{}c@{}} ``A cubism painting \\ of  \pholdercolor" \end{tabular}} & {\begin{tabular}{c@{}c@{}c@{}c@{}} ``A \pholdercolor{} as a knight \\ in plate armor" \end{tabular}} \\
    \end{tabular}
    \end{tabular}}
    \caption{Out of domain personalization results. Our method enables personalization of nearby domains (dogs with a model trained on cats, human sketches with a model trained on faces), but it fails when moving further away. Note that succesfully out-of-domain inversions typically require more fine-tuning steps.
    }
    \label{fig:cross_domain} 
\end{figure}

While our method can achieve high-fidelity personalization with short training times, it is not free of limitations. First, our encoders learn to generalize from large datasets that represent the coarse target class. As such, they are only applicable for classes where large datasets exist. In practice, this includes both faces and artistic styles which are the current primary use-cases for personalization. However, it may limit their applicability to rare, one-of-a-kind objects. In \cref{fig:cross_domain} we show the effects of trying to personalize out-of-domain images using our method. When the concepts are from nearby domains (dogs, inverted with a model trained on cats), the method can still produce high-fidelity results. For farther domains (a wooden toy) the method fails to capture concept-specific details.

A further limitation is in the need to perform inference-time tuning. While the impact on synthesis times is short, our approach does require the inference-machine to be capable of tuning a model. Moreover, as the encoder and text-to-image models must be tuned in tandem, this process requires more memory than direct fine-tuning approaches.

\section{Conclusions and Future Work}\vspace{3pt}

We introduced an encoder-based domain-tuning method for fast personalization of text-to-image models.
At the core of our method is the idea that large, domain-specific datasets can be leveraged to find a good starting point for future optimization, thus allowing the network to better adapt to novel samples from the same domain. In this sense, our work draws inspiration from meta learning methods. As our results demonstrate, large text-to-image models are amenable to such approaches even without resorting to their typically complex machinery. Importantly, our approach allows us to achieve remarkable acceleration while maintaining state-of-the-art quality.

In the future, we plan to further investigate encoder-based personalization methods, with a focus on improving the HyperNetwork based approach. We believe that with the proper regularization, this approach can be improved, leading to instant, training-free personalization.

\section{Ethic statement}
Text-to-image models may be used to create misleading content or promote disinformation. Single-image personalization may increase the ability to forge convincing images of non-public individuals.

Text-to-image models are susceptible to biases found in the training data. Our work builds on such models and may exhibit and be used to propagate similar biases. However, as demonstrated in \citep{gal2022image}, personalization can also be used to reduce model biases.

Finally, the ability to learn artistic styles may be misused for copyright infringement. However, recent work~\citep{shan2023glaze} has shown that it is possible to protect artwork from being copied by text-to-image generators, and we hope that future research in this direction could serve to mitigate such risks of infringement.
\begin{acks}
This work was partially supported by Len Blavatnik and the Blavatnik family foundation, the Deutsch Foundation, the Yandex Initiative in Machine Learning, BSF (grant 2020280) and ISF (grants 2492/20 and 3441/21). 
\end{acks}

\begin{figure*}
    \centering
    \includegraphics[width=.96\textwidth]{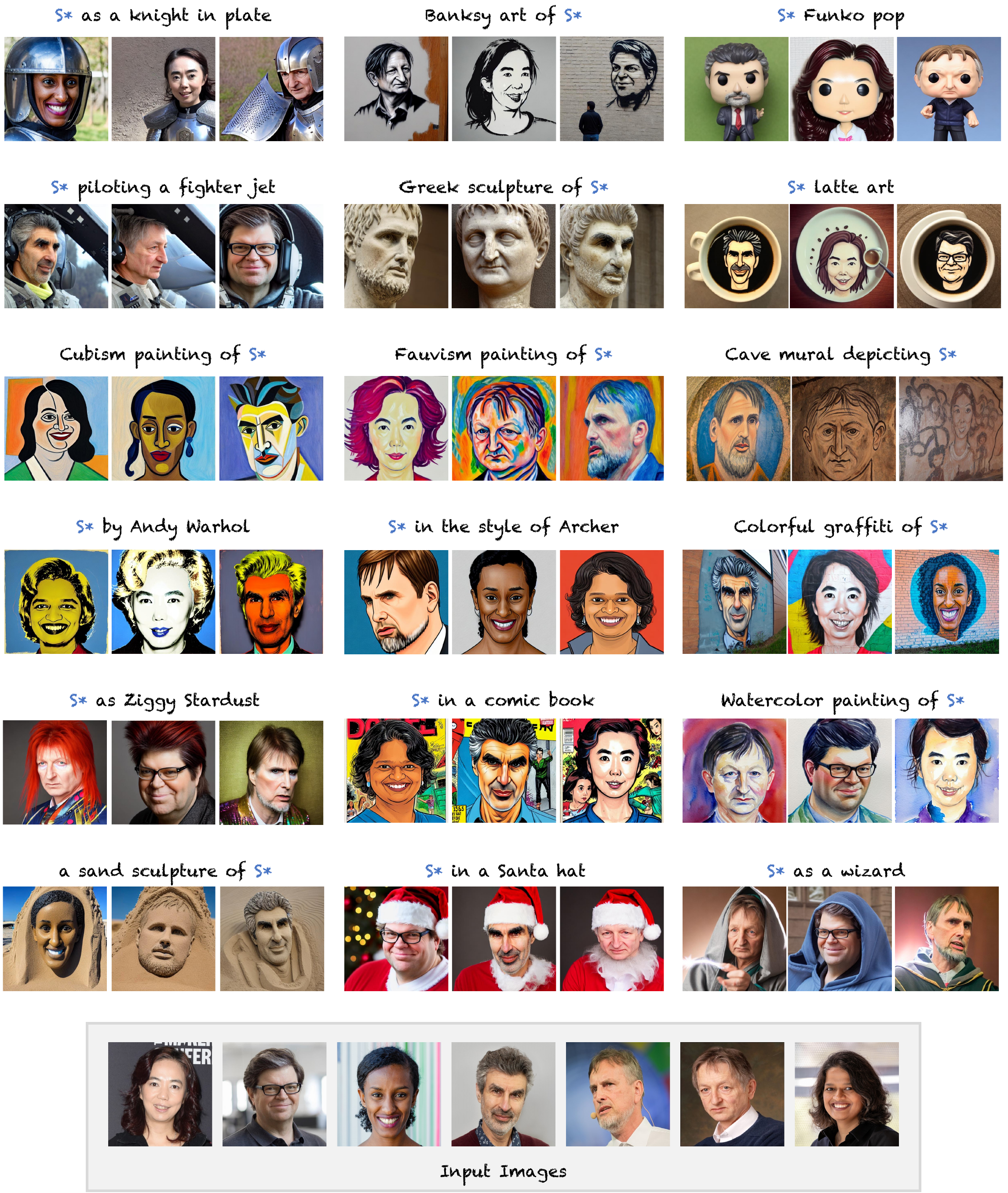}
    \caption{Additional personalized synthesis results created using our method. For each identity, the single input image is shown below.}
    \label{fig:additional_results}
\end{figure*}

\bibliographystyle{ACM-Reference-Format}
\bibliography{main}

\end{document}